\documentclass{article} 
\usepackage{iclr2026_conference,times}

\usepackage{amsmath,amsfonts,bm}

\def\eqref#1{equation~\ref{#1}}

\def\1{\bm{1}}

\DeclareMathAlphabet{\mathsfit}{\encodingdefault}{\sfdefault}{m}{sl}
\SetMathAlphabet{\mathsfit}{bold}{\encodingdefault}{\sfdefault}{bx}{n}

\usepackage{hyperref}
\usepackage{url}
\usepackage[utf8]{inputenc} 
\usepackage[T1]{fontenc}    
\usepackage{hyperref}       
\usepackage{url}            
\usepackage{booktabs}       
\usepackage{amsfonts}       
\usepackage{nicefrac}       
\usepackage{microtype}      
\usepackage{xcolor}  
\usepackage{subcaption}
\usepackage{booktabs}
\usepackage{multirow}
\usepackage{graphicx}
\usepackage{wrapfig}
\usepackage{enumitem}
\usepackage[ruled,vlined]{algorithm2e}
\usepackage{float}
\usepackage{amsmath}       
\usepackage{amssymb}       
\usepackage{mathtools}
\usepackage{wrapfig}
\newcommand{\algo}{\textsc{SRT}}

\usepackage[most,skins,theorems]{tcolorbox}
\definecolor{rliableolive}{HTML}{BBCC33}
\tcbset{
  aibox/.style={
    width=\linewidth,
    top=10pt,
    bottom=4pt,
    colback=blue!6!white,
    colframe=black,
    colbacktitle=black,
    enhanced,
    center,
    attach boxed title to top left={yshift=-0.1in,xshift=0.15in},
    boxed title style={boxrule=0pt,colframe=white,},
  }
}
\newtcolorbox{AIbox}[2][]{aibox,title=#2,colback=rliableolive!10!white,#1}
\usepackage[most]{tcolorbox}

\tcbset{
  examplebox/.style={
    width=\linewidth,
    colback=gray!5,
    colframe=black,
    coltitle=black,
    fonttitle=\bfseries,
    boxrule=0.4pt,
    arc=3pt,
    left=6pt,
    right=6pt,
    top=6pt,
    bottom=6pt,
    enhanced
  }
}

\title{Can Large Reasoning Models Self-Train?}

\author{Sheikh Shafayat$^{1}$\thanks{Equal contribution.
  \\ \indent Correspondence: \texttt{sheikh.shafayat@kaist.ac.kr}, \texttt{ftajwar@andrew.cmu.edu}}
  \quad
  Fahim Tajwar$^{2}$\footnotemark[1]
  \\[0.5ex]
  \textbf{Ruslan Salakhutdinov}$^{2}$
  \quad
  \textbf{Jeff Schneider}$^{2}$
  \quad
  \textbf{Andrea Zanette}$^{2}$
  \\[1ex]
  $^{1}$KAIST \quad $^{2}$Carnegie Mellon University
}

\iclrfinalcopy 
\begin{document}

\maketitle

\begin{abstract}

Recent successes of reinforcement learning (RL) in training large reasoning models motivate the question of whether self-training — the process where a model learns from its own judgments — can be sustained within RL. In this work, we study this question using majority voting as a simple self-feedback mechanism. On a comprehensive set of experiments on both synthetic and real reasoning tasks, we find that this basic approach improves not only the model's reasoning performance, but also its capability of generating better quality feedback for the next RL iteration, driving further model improvement. Yet our analysis also reveals a critical limitation of such a self-training paradigm --- prolonged RL with self-reward leads to reward hacking where models learn to maximize training (pseudo-)reward,
resulting in sudden and complete performance collapse. 
Together, these results highlight feedback design as the central challenge and call for future research on mechanisms to enable prolonged self-improvement.

\end{abstract}

\section{Introduction}

Pre-training on human-curated corpora has endowed language models with broad general-purpose capabilities \citep{brown2020language,rae2022scalinglanguagemodelsmethods}, but the supply of such data is becoming a bottleneck as compute scales rapidly \citep{hoffmann2022training,sevilla2022compute}. Reinforcement learning (RL)~\citep{sutton1998reinforcement} with verifiable rewards (RLVR) addresses this limitation by using automatic correctness checks, and has already shown success in reasoning and agentic tasks \citep{deepseekai2025deepseekr1incentivizingreasoningcapability,openai2024openaio1card}. Yet, achieving super-intelligence inevitably requires moving beyond domains where humans can provide ground-truth solutions. In these domains, an appealing approach is to have models \textbf{self-improve}— judging the correctness of their own outputs and using this signal to refine future generations \citep{zelikman2022star,song2025mind,huang2025selfimprovement}. If sustained iteratively, and \textbf{if teacher models can also self-improve}, this process could enable continual progress without human supervision — which is crucial step for AI based automatic scientific discovery.

Prior work on self-improvement has mainly relied on supervised fine-tuning (SFT) \citep{zelikman2022star,huang-etal-2023-large} or direct preference optimization (DPO) \citep{rafailov2024directpreferenceoptimizationlanguage,prasad2024selfconsistencypreferenceoptimization}, where the self-labeling rule is updated only a handful of times (e.g., 1–10 rounds) and \emph{is generally kept fixed within a training round}. These studies show that self-improvement \emph{can} be effective, but leave open whether it can be sustained over longer horizons. Moreover, they are fundamentally bounded by the verification capabilities of the teacher model used to obtain training supervision, which is kept fixed for extended periods of training. In contrast, RL continuously updates the model on freshly generated (``on-policy'') data~\citep{xu2024dposuperiorppollm,tajwar2024preferencefinetuningllmsleverage,lanchantin2025bridgingofflineonlinereinforcement}, a property that has been critical to its success in training reasoning models with verifiable rewards. This success raises a natural question: can self-improvement leverage the same continuous-update paradigm? To study this question, we investigate the setting in which the feedback signal is \textbf{updated at every gradient step}—fundamentally altering the dynamics of self-improvement compared to earlier work.

In this RL-based setup, the choice of feedback mechanism is critical. If feedback is always correct (e.g., perfectly verifying mathematical solutions), the procedure reduces to standard RL with ground-truth supervision. However, in practice, self-feedback is imperfect, and its design determines whether self-training is effective or not. As a first step, we study the simplest possible signal: \emph{majority vote}.~\citet{wang2023selfconsistency} has empirically demonstrated that majority vote tends to have higher accuracy compared to individual generations. Here, we cast the majority vote mechanism as a reward function --- granting positive reward to model outputs that match the most common answer.
Previous works~\citep{huang-etal-2023-large,prasad2024selfconsistencypreferenceoptimization} have employed majority voting mainly as a mechanism to extract better quality generations from a \emph{fixed} teacher policy to then distill it into the student model. 
\citet{zuo2025ttrl} concurrently  examined a similar RL procedure with majority voting, but in a different setting, as their focus is on training and testing on the same set of prompts. In contrast, our aim is to use this simple pseudo-label generation mechanism to investigate the validity of RL powered self-training frameworks. 

Our key contributions are threefold:
\begin{enumerate}[label=\textbf{(\arabic*)}]
    \item Our comprehensive set of experiments demonstrate that this simple mechanism yields measurable gains over the base model on key reasoning metrics such as maj@k and avg@k success rates (Section~\ref{sec:question_1}). Remarkably, we observe clear improvement in the label generating policy after each gradient step, and this translates to gains over employing labels from a fixed teacher. Moreover, self-training achieves comparable performance to RLVR on 4 different base models.

    \item In synthetic tasks where one can control the difficulty of the training dataset, we observe that a simple curriculum-based self-training approach can enable the model to keep climbing on progressively harder tasks without ground-truth labels (Section~\ref{sec:question_2}). Overall, this signals a promising path towards self-improvement of LLMs.

    \item However, prolonged training with this framework consistently teaches models to ignore the prompt entirely and output the same template final answer, which maximizes training reward but leads to complete collapse on test datasets. To our knowledge, we are the first to demonstrate this failure mode of self-improving LLMs by employing self-consistency based training signal. In this work, we analyze the dynamics of model collapse in depth and trace them to the self-reinforcing nature of imperfect feedback.
\end{enumerate}

Overall, these findings identify feedback design as the key challenge that future research should address to sustain self-improvement.

Our code is publicly available here: \url{https://github.com/tajwarfahim/srt}. For other project-related assets such as constructed datasets, please see our project website at \url{https://self-rewarding-llm-training.github.io/}.

\begin{figure}[t!]
    \centering
    \includegraphics[width=0.98\linewidth]{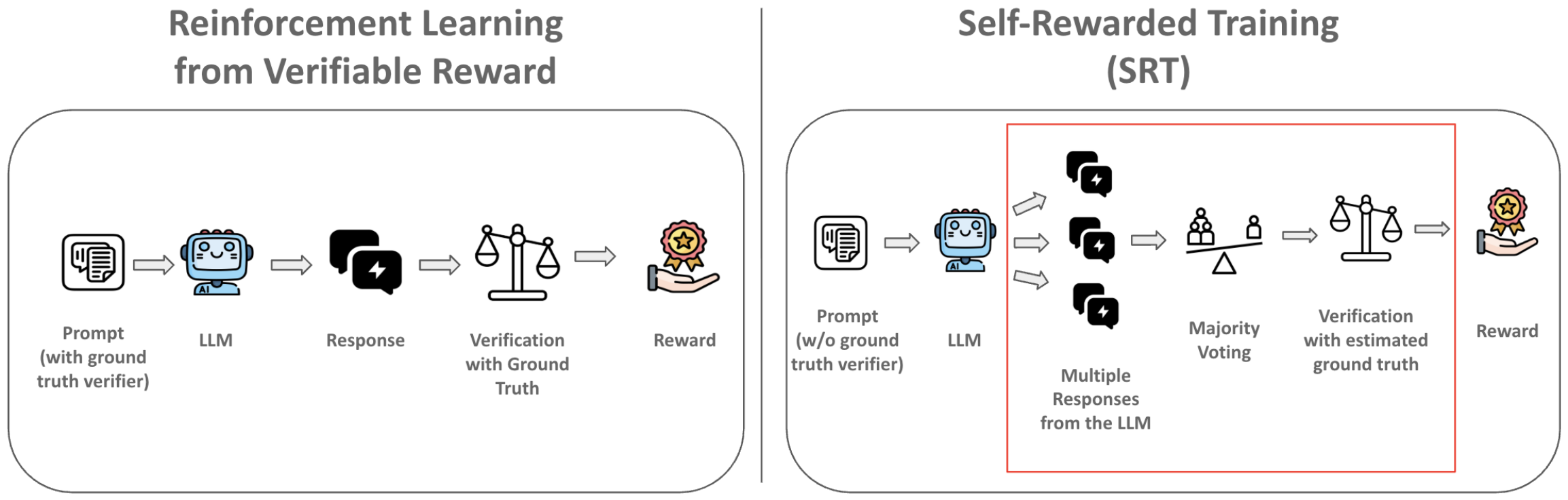}
    \caption{\footnotesize{\textbf{(Overview of \algo{})} In RLVR, one produces the reward for RL training using a ground truth verifier. Contrary to that, \algo{} does not assume access to a ground truth verifier; instead it uses majority voting from the model's own generations to estimate the ground truth, and use this proxy reward signal to train the model.}}
    \label{fig:teaser_figure}
    \vspace{-0.5cm}
\end{figure}

\def\policy{\pi_\theta}
\def\reward{r}
\def\promptDist{\mathcal X}
\def\advantage{\mathcal A}
\newcommand{\prompt}{x}
\newcommand{\response}{y}
\newcommand{\trueanswer}{y^*}
\newcommand{\genacc}{\text{Acc}_\text{gen}(\theta)}
\newcommand{\veracc}{\text{Acc}_\text{ver}(\theta,n)}
\newcommand{\verify}{\text{Verify}}
\newcommand{\verifier}{f}
\newcommand{\rewardver}{R_v}
\newcommand{\rpos}{r_{\text{pos}}}
\newcommand{\rneg}{r_{\text{neg}}}

\section{Preliminaries}

Let $\policy$ denote a language model parameterized by $\theta$. Given a prompt $\prompt$, the model produces a response $\response = (\response^{1}, \response^{2}, \dots)$ auto-regressively. Formally, each token in the response sequence is generated according to the conditional probability:
\begin{equation}
    \response^{k+1} \sim \policy(\cdot \mid \prompt, \response^{\leq k}),
\end{equation}
where we use $\response^{\leq k}$ to refer to the first $k$ tokens generated by the model.

For reasoning-based tasks considered here, the model typically produces responses following a step-by-step ``chain-of-thought'' reasoning approach \citep{wei2022chain}. A verification function $\reward(\response)$, whose dependence on $\prompt$ is omitted for brevity, extracts the model's proposed solution from the generated response and evaluates its correctness against the prompt-specific ground-truth answer:
\begin{equation}\label{eqn.def.reward.binary}
    r(\response) = \begin{dcases}
        1 & \text{if $\response$ is correct,}\\
        0 & \text{if $\response$ is incorrect.}
    \end{dcases}
\end{equation}

Typically, one optimizes the expected \emph{pass rate}, defined as the average accuracy across a distribution of prompts $\promptDist$: 
\begin{equation}\label{eqn.objective}
    J(\theta) = \mathbb{E}_{\prompt \sim \promptDist}\mathbb{E}_{\response \sim \policy(\cdot \mid \prompt)}[\reward(\response)].
\end{equation}

Taking the gradient of the objective function (\ref{eqn.objective}) with respect to $\theta$ and employing a baseline for variance reduction \citep{sutton1998reinforcement} leads to the well-known policy gradient formulation:
\begin{equation}\label{eqn.policy.gradient}
\nabla_{\theta} J(\theta) = \mathbb{E}_{\prompt \sim \promptDist}\mathbb{E}_{\response \sim \policy(\cdot \mid \prompt)}\left[\advantage(\response)\nabla_{\theta}\log \policy(\response \mid \prompt)\right],
\end{equation}
where the advantage function $\advantage(\response)$ is given by:
\begin{equation}\label{eqn.advantage}
\advantage(\response) = \reward(\response) - \mathbb{E}_{\response' \sim \policy(\cdot \mid \prompt)}[\reward(\response')].
\end{equation}
Here $\mathbb{E}_{\response' \sim \policy(\cdot \mid \prompt)}[\reward(\response')]$ is the average pass rate for prompt $\prompt$.
In practice, the policy gradient in ~\eqref{eqn.policy.gradient} is estimated through Monte Carlo samples, yielding the classical REINFORCE algorithm \citep{williams1992simple}. Recent works have modified this base policy gradient formulation to improve its stability, efficiency, and practicality, resulting in advanced methods such as REINFORCE++ \citep{hu2025reinforceefficientrlhfalgorithm}, GRPO \citep{shao2024deepseekmathpushinglimitsmathematical}, PPO \citep{schulman2017proximal}, RLOO \citep{ahmadian-etal-2024-back}, Dr. GRPO~\citep{liu2025understandingr1zeroliketrainingcritical}, GSPO~\citep{zheng2025groupsequencepolicyoptimization}, etc. Implicit to the training method used in our work is the notion of a generation-verification gap \citep{song2025mind}, where generating correct solutions is hard, but verifying them is easy. We present its definition in Appendix \ref{app:gen-ver-gap}.

\section{Self-Rewarded Training}

\begin{wrapfigure}{r}{0.6\textwidth}
\begin{minipage}{0.6\textwidth}
\begin{algorithm}[H]
\caption{Self-Rewarded Training (\algo{})}
\label{alg:srt}
\KwIn{Prompt dataset $\mathcal{X}$}
\ForEach{RL iteration}{
    \tcc{\emph{Inference step}}
    Sample minibatch $\mathcal{B} \subseteq \mathcal{X}$

    \ForEach{prompt $x \in \mathcal{B}$}{
        Generate $n$ solutions $y^{(1)},...,y^{(n)} \sim \pi_\text{label}(\cdot|x)$

        Identify majority-vote answer:
        $$y_\text{majority} \gets \arg\max_{y'} \sum_{i=1}^{n} \mathbf{1}[\text{answer}(y^{(i)}) = y']$$

        Define reward function: $$r(y) \gets \mathbf{1}[\text{answer}(y) = y_\text{majority}]$$
    }

    \tcc{\emph{Gradient update step}}
    Perform RL gradient update using $r(\cdot)$
}
\end{algorithm}
\end{minipage}
\vspace{-1em} 
\end{wrapfigure}

Our objective in this work is to investigate whether \emph{reliable} training supervision for language models can be generated without external labels. The typical practice for online RL involves generating multiple responses to a prompt, then assigning high or low rewards to each generation according to a ground-truth verifier. In the absence of such a verifier, one might develop a mechanism to derive proxy labels. This mechanism then provides a simple recipe for framing self-improvement as an RL problem. At a high level, each iteration proceeds as follows: \textbf{(1)} Sample a mini-batch of prompts, \textbf{(2)} Determine pseudo-labels, $y_\text{pseudo}$, using the mechanism for each prompt, \textbf{(3)}
Generate $n$ responses per prompt and use agreement with the derived psuedo-labels as an intrinsic binary reward:

\begin{align}
    \label{eqn:self-reward}
    r(y)=\mathbf{1}[\text{answer}(y)=y_\text{pseudo}],
\end{align}

and then \textbf{(4)} Perform a single RL update step on this mini-batch using the  reward function $r(\cdot)$.

\paragraph{Self-supervision via majority voting.}
Among several possible choices for determining reasonably accurate pseudo-labels, we explore the simplest one in this work, \emph{majority voting}. Majority voting has been empirically demonstrated to have higher accuracy compared to individual model generations~\citep{wang2023selfconsistency} and is thus a suitable choice to exploit an LLM's inherent generation-verification gap (see Appendix Figure~\ref{fig:generation_verification_gap}, where majority voting consistently outperforms the model's average capabilities across 3 test datasets and therefore can act as a verifier). 
In our setting, models typically produce a step-by-step chain of thought followed by a final answer (in the case of regular RL training, this final answer is extracted and matched with the ground truth to produce rewards), so one can group all responses by their final answer to determine the majority vote. Concretely, assume we want to generate pseudo-labels using policy $\pi_\text{label}$ (which can be any reasonable policy). This procedure then involves: \textbf{(1)} sampling multiple answers per prompt using policy $\pi_\text{label}$, \textbf{(2)} grouping answers according to their parsed final solutions, \textbf{(3)} estimating the ground truth answer with the most common solution.

\paragraph{Self-Rewarded Training (SRT).} 
The general procedure is described in Algorithm~\ref{alg:srt}, which henceforth shall be called Self-Rewarded Training (\algo{}) in this paper. Since the method prescribes a specific form of the reward function using model self-consistency, it is compatible with all the common RL training algorithms such as PPO, RLOO, REINFORCE, GRPO, etc. We study the quality of generated labels during training by controlling $\pi_\text{label}$: setting $\pi_\text{label}$ to be the base model recovers our familar setting of learning the majority voting decisions of a fixed model (while still using the current policy's rollouts for RL training), and setting $\pi_\text{label}$ to be the current policy $\pi_\theta$ after each gradient step allows us to study whether the quality of the learning signal can be concurrently improved during RL training. In the case where we use the current policy $\pi_\theta$ to generate our pseudo-labels, we can reuse them for performing the RL gradient step as well. As the number of generations per prompt typically falls in the range 16-64~\citep{yu2025dapoopensourcellmreinforcement}, this variant of \algo{} incurs no additional compute cost compared to the versions of these algorithms employing ground truth labels. In our work, whether the final answer is correctly formatted and parseable is used to filter responses, but given more compute, one can theoretically employ more sophisticated systems like LLM-as-a-judge~\citep{zheng2023judgingllmasajudgemtbenchchatbot,gu2025surveyllmasajudge} or generative verifiers~\citep{zhang2025generativeverifiersrewardmodeling} to further improve the quality of the training signal. We leave these for future work.

As long as majority voting leads to a positive generation-verification gap at each RL iteration, we expect iterative self-rewarding to provide a useful supervisory signal. We describe our empirical observations in the following section.

\section{Experiments and Analysis}
\label{sec.exp}

\begin{figure}[t]
    \centering
    \includegraphics[width=0.95\linewidth]{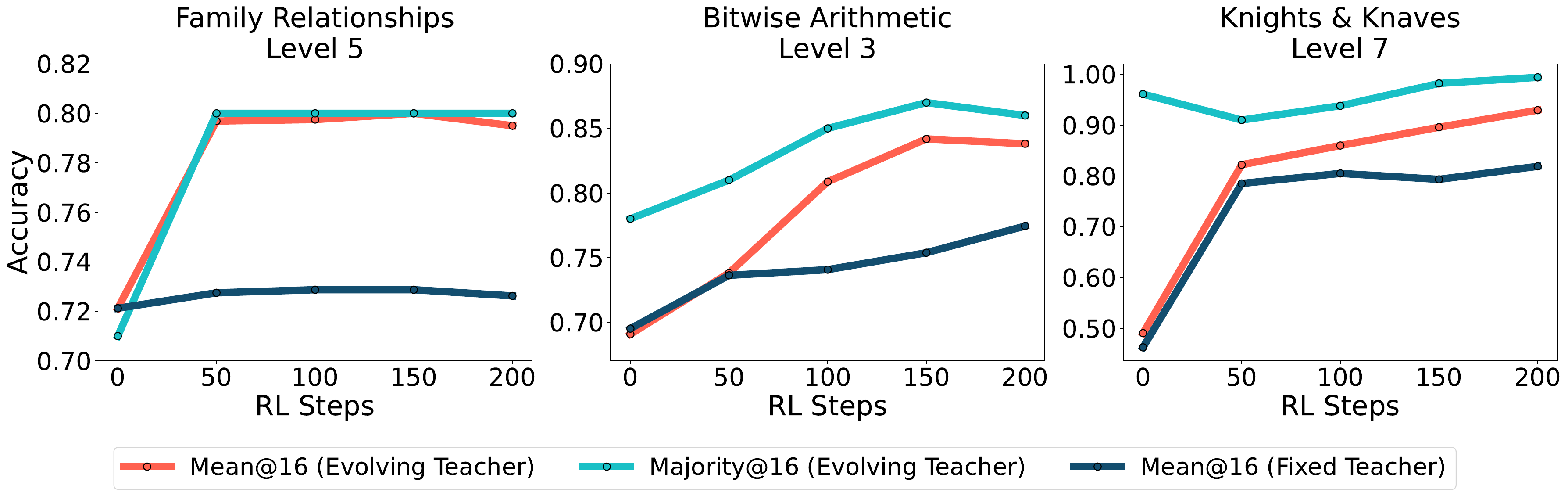}
    \caption{\footnotesize (\textbf{\algo{} improves both performance and quality of generated labels during training.}) We investigate self-training under controlled settings on synthetic reasoning tasks from Reasoning Gym. Remarkably, \algo{} improves not only the mean accuracy, but the majority voting accuracy as well, which is the source of our training supervision. Improvement in the quality of training signal drives further improvement in performance, as \algo{} outperforms its variant employing the majority votes from a fixed teacher as proxy labels.}
    \label{fig:reasoning-gym-main-plot}
\end{figure}

In this section, we present the results of our empirical study. Our primary aim is to answer the two following research questions: \textbf{(1)} \textbf{Can \algo{} improve an LLMs' reasoning abilities beyond the base model's capabilities}, 
in terms of both the quality of training reward signal and performance on downstream tasks? Specifically, since simply performing majority voting on the base model can improve performance compared to individual generations, can we improve the quality of the majority votes (which acts as the verifier for \algo{}) in addition to pass@1 performance compared to the base model via \algo{}?  \textbf{(2)} If \algo{} can go beyond the base model, \textbf{can this improvement be sustained indefinitely?} We systematically design experiments to answer these questions below.

\subsection{Can \algo{} go beyond the base model's capabilities?} \label{sec:question_1}

There are two potential axes of improvements over the base model for \algo{} that can be studied: (1) the improvement in accuracy over held-out prompts, and (2) the improvement in the quality of generated labels themselves during the training procedure. Unlike previous works~\citep{huang-etal-2023-large,prasad2024selfconsistencypreferenceoptimization} that distill the majority voting decision of a fixed policy (typically the base model) into the current model, we want to study \textbf{whether the quality of the majority votes of the evolving policy improves} as a result of self-improvement. 

\begin{figure}[t]
    \centering
    \includegraphics[width=0.99\linewidth]{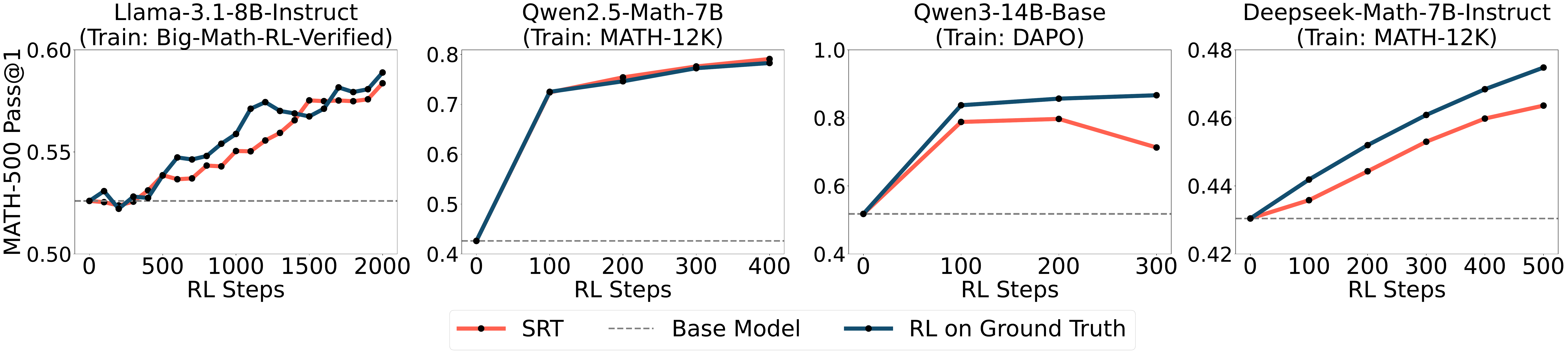}
    \caption{\footnotesize \textbf{(Evaluating \algo{} on real-world math problems.)} Comparison between \algo{} and RL with ground truth across different base models and training datasets. Following~\citep{Oertell2024Heuristics}, all models are trained using RLOO (for experiments with GRPO, see Figure~\ref{fig:grpo_vs_rloo_all_test_datasets}) and tested using average pass@1 accuracy on MATH-500. \algo{} achieves comparable performance to that of ground-truth training across different base models. For training curves using more combinations of (train, test) dataset pairs, refer to Appendix~\ref{app:qwen2.5_math_7b_full_results} and~\ref{app:qwen3_full_results}.}
    \label{fig:all_model_average_accuracy_math_500}
\end{figure}

\begin{figure}[t]
    \centering
    \includegraphics[width=0.99\linewidth]{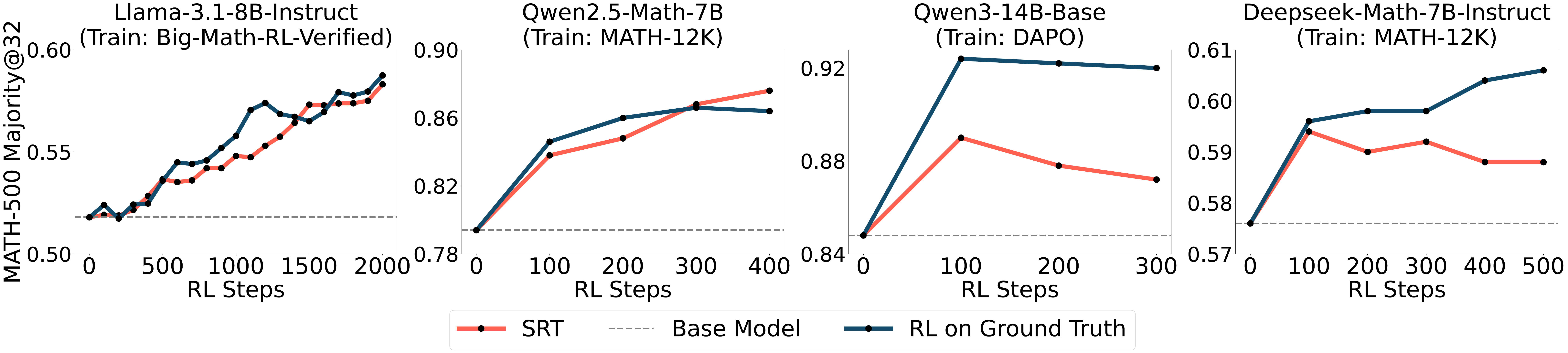}
    \caption{\footnotesize \textbf{(Majority@32 accuracy comparison between \algo{} and RL with ground truth)} We compare the majority@32 accuracy, as opposed to average accuracy shown in Figure~\ref{fig:all_model_average_accuracy_math_500}. \textbf{Note that for Llama-3.1-8B-Instruct, we use the official model card evaluation temperature of 0, hence majority@32 is the same as average@32 accuracy.} \algo{} shows improvement in the quality of the majority votes themselves, which distinguishes our algorithm from that of learning from a fixed teacher's majority votes.}
    \label{fig:all_model_majority_voting_accuracy_math_500}
\end{figure}

\paragraph{Experiments on synthetic reasoning tasks.} 
Complex reasoning tasks like math require domain-specific pre-training/midtraining for online RL to be effective~\citep{wu2025miragemethodmodeltaskalignment,gandhi2025cognitivebehaviorsenableselfimproving}; it is also more difficult to control the difficulty of the individual tasks that the model sees during training. Therefore, we first study this question using synthetic reasoning tasks from REASONING GYM~\citep{stojanovski2025reasoninggymreasoningenvironments}, a collection of over 100 reasoning environments for reinforcement learning with verifiable rewards. These tasks can also be generated with adjustable difficulty, making it a suitable test-bed for our work. Concretely, we use 3 tasks from Reasoning Gym: \textbf{(1)} \textbf{Family Relationships}, a logic puzzle involving a group of individuals connected via different relationships, and the model has to reason about the relationship between two individuals within this group, \textbf{(2)} \textbf{Bitwise Arithmetic}, a task for testing models' understanding of Bitwise Arithmetic operations, and \textbf{(3)} \textbf{Knights \& Knaves}, a logic puzzle involving characters who always either tell the truth (knights) or always lie (knaves), and the challenge is to deduce who is who based on their statements. Appendix~\ref{app:reasoning_gym_example_tasks} shows example prompts from each of these tasks. We use a Qwen-3-4B-Base~\citep{yang2025qwen3technicalreport} model for all our reasoning gym experiments.

Since our goal is to study whether \algo{} can improve performance on top of a reasonably strong base model, following the setting of~\citet{lee2025selfimprovingtransformersovercomeeasytohard}, we first train the base model with GRPO using ground truth labels on the easiest difficulty setting. This is used to teach the base model proper formatting rules and how to solve the basic task before we train using \algo{} on the next level of difficulty without ground-truth labels. The detailed training settings can be found in Appendix~\ref{app:training_hparam_details}.

\paragraph{Results.} Figure \ref{fig:reasoning-gym-main-plot} shows our main results: \algo{} using the current policy as the label generator improves both avg@16 and majority vote@16 accuracy on all 3 reasoning gym tasks. Since we derive our training signal from the majority votes of the current policy evolving with every RL step --- this demonstrates that self-improvement using \algo{} can progressively improve the quality of the pseudo-labels as well. Note that this would not be possible in prior works~\citep{prasad2024selfconsistencypreferenceoptimization,huang-etal-2023-large} which distills the majority votes of a \emph{fixed teacher} policy throughout one round of training. We expect the improvement in the \emph{evolving teacher} policy to result in further performance gain. To validate this, we compare \algo{} with a variant of the same algorithm where we use the majority votes of the fixed starting policy instead of the evolving current policy as pseudo-labels. Figure \ref{fig:reasoning-gym-main-plot} shows their comparison: in Family Relationships and Bitwise Arithmetic, we see larger gains in majority voting performance, and likewise \algo{} outperforms its fixed teacher variant substantially, by 10\% for Bitwise Arighmatics, 8\% in Family Relationship, and 6\% in Knights and Knaves. On Knights \& Knaves, the starting policy already has >90\% majority voting accuracy, and we see the difference between the evolving teacher and fixed teacher variants of \algo{} to be smaller here. Furthermore, these performance gains cannot be explained by learning to format properly: since the model was trained on an easier difficulty level, it can already format its responses correctly in most cases (Appendix~\ref{app:parsability_metrics}), and we see performance climbing after saturation in model's format correctness. Overall, on the synthetic reasoning tasks, \textbf{\algo{} clearly pushes the model beyond its starting capabilities}, showing the promise of self-improvement even from this basic recipe for self-supervision.

We show a summary of our empirical findings using different base models and training datasets in Figures~\ref{fig:all_model_average_accuracy_math_500} and~\ref{fig:all_model_majority_voting_accuracy_math_500} (for training curves using more combinations of (train, test) dataset pairs, refer to Appendix~\ref{app:qwen2.5_math_7b_full_results} and~\ref{app:qwen3_full_results}). On all instances, \algo{} improves both avg@16 and majority vote@16 accuracies on heldout MATH-500 prompts, and performs on par with regular RL training with ground truth verification. More impressively, the observations hold for base models like Llama-3.1-8B-Instruct, which is known to be particularly difficult for RL training on reasoning tasks~\citep{gandhi2025cognitivebehaviorsenableselfimproving}, improving its average accuracy from 52.6\% to nearly 60\%.

We also compare \algo{} with its offline variants: SFT on the majority vote~\citep{huang-etal-2023-large}, DPO and ScPO~\citep{prasad2024selfconsistencypreferenceoptimization} employing contrastive learning between the majority vote and non-majority vote answer in Table~\ref{tab:baseline_comparison}. We observe that \algo{} retains better performance compared to its offline variants distilling the majority vote decisions of the base policy, \textbf{showing the benefit of self-improvement in the label-generating policy}. For more details about the baselines, including choice of hyperparameters and performance on additional test datasets, refer to Appendix~\ref{app:baseline_details}.

\begin{table}[h!]
    \centering

    \caption{\footnotesize \textbf{(Peak Performance)} We compare \algo{} against various baselines using a Qwen2.5-Math-7B model. Reported numbers are peak mean@32 accuracy, averaged over 3 heldout datasets: AIME 2024, AIME 2025, AMC. We find that \algo{} generally retains better performance compared to baseline methods distilling the majority voting decisions of the fixed teacher policy into the current model.}
    
    \resizebox{0.75\textwidth}{!}{%
        \begin{tabular}{c|ccccc|c}
        \toprule
            Train Data & Base Model & SFT & DPO & ScPO & \algo{}  & RL on Ground Truth \\
        \midrule
        MATH-12K & 0.15 & 0.18 & 0.23 & 0.20 & 0.32 & 0.33 \\
        DAPO & 0.15 & 0.18 & 0.21 & 0.20 & 0.31 & 0.36\\
        \bottomrule
        \end{tabular}
    }
    \vspace{10pt}
    \label{tab:baseline_comparison}
\end{table}

\begin{AIbox}{Takeaway 1: \algo{} can improve reasoning capabilities beyond the base model}
On both synthetic and real reasoning tasks, \algo{} improves average and majority voting accuracies, showing ability gains beyond the base model. Specially, improvement in majority voting accuracy also signifies improvement in the quality of generated self-supervision during training, demonstrating the importance of training the ``verifier'' in addition to just training the policy. Overall, our simple recipe shows a promising path forward to self-improvement. 
\end{AIbox}
\vspace{-0.1cm}

\subsection{Can Self-Improvement from \algo{} be Sustained Indefinitely?} \label{sec:question_2}

Given the strong performance of \algo{} in various reasoning tasks and model architectures, an important question is whether self-training can be maintained over extended iterations. Similar to the prior section, we first test \algo{} on synthetic tasks with controllable difficulty to rigorously study its properties, and then test the resulting insights on real-world reasoning domains.

\begin{figure}[h]
    \centering
    \vspace{-0.2cm}
    \includegraphics[width=0.99\linewidth]{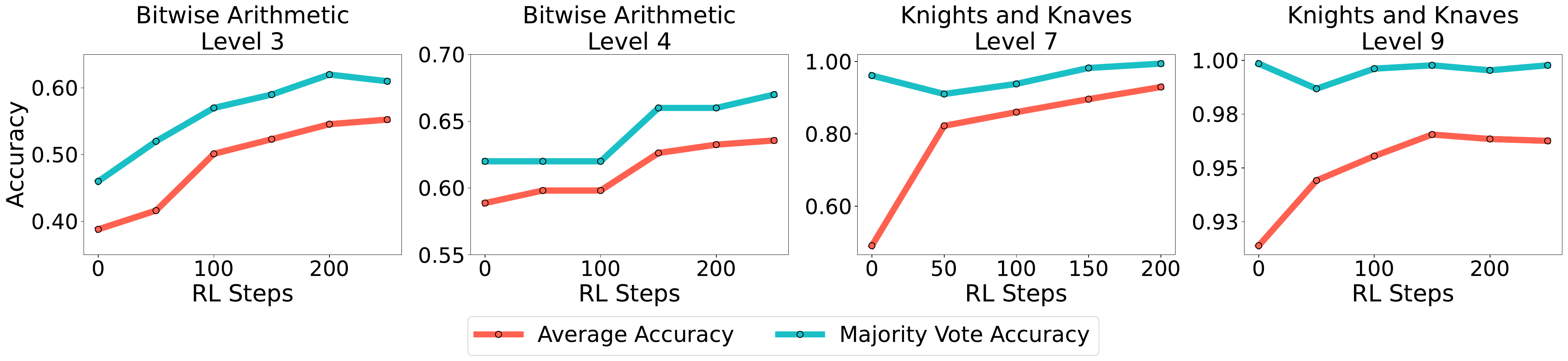}
    \caption{\footnotesize \textbf{(Multi-level climbing on Reasoning Gym using curriculum)} The Qwen3-4B-Base model can climb on progressively more difficult tasks without ground truth labels via a simple curriculum strategy --- where we train an earlier level's final checkpoint with \algo{} on the next difficulty level. This approach also seems to improve both average and majority voting accuracy on each level.}
    \label{fig:multi-step-climbing}
    \vspace{-0.4cm}
\end{figure}

\paragraph{Multi-level self-improvement on synthetic tasks using curriculum.} Since Reasoning Gym provides a built-in way to control the difficulty of generated tasks, we first investigate whether self-training on an easier set of tasks can produce a model capable of self-improvement on progressively harder levels of difficulty. To do so, we choose 2 Reasoning Gym tasks: Bitwise Arithmetic and Knights \& Knaves. Similar to our previous setting, we first train using RL on ground truth labels on the easiest level of difficulty, then progressively train on harder levels without ground truth labels (i.e., \algo{} on level 5 starts with the checkpoint obtained from \algo{} on level 4, and so on). For more details, refer to Appendix~\ref{app:training_hparam_details}. 

Figure \ref{fig:multi-step-climbing} shows our primary results: in this controlled setting, \algo{} is able to maintain self-improvement on progressively harder difficulties. In particular, \algo{} can show reasonable improvement in Bitwise Arithmetic Level 4 after being initialized on Level 2 with ground-truth training, and also progressively climb to near 100\% accuracy on Knights \& Knave Level 9 after being trained with ground-truth on Level 2 only (intermediate levels are trained with \algo{}). In synthetic reasoning tasks where we have a knob for controlling the difficulty of training problems, even the simple curriculum of progressively training on more difficult problems with self-supervision can maintain improvement across multiple difficulty levels.   

\begin{figure}[h]
    \centering
    \includegraphics[width=0.99\linewidth]{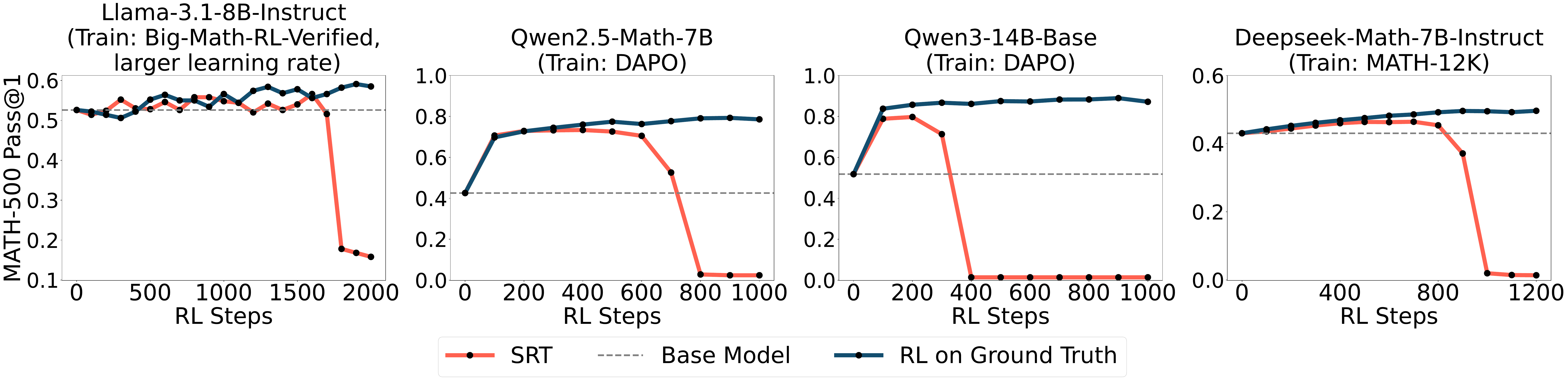}
    \caption{\footnotesize \textbf{(Extended self-training leads to model collapse)} Inspired by multi-level improvement on reasoning gym tasks, we take four LLMs with strong math abilities from pretraining, and train them with \algo{} for an extended period of time. \algo{} improves performance at first, but then demonstrates complete model collapse on all 4 base models. (Note: on Llama-3.1-8B-Instruct, the learning rate used in Figures~\ref{fig:all_model_average_accuracy_math_500} and ~\ref{fig:all_model_majority_voting_accuracy_math_500}, $10^{-7}$, does not lead to model collapse within our training budget, but $3 \times 10^{-7}$, a slightly higher learning rate does --- we hypothesize that with an extended training run, even $10^{-7}$ would lead to model collapse. For more details on the effect of hyperparameters on model collapse, refer to Figure~\ref{fig:training_choice_summary}.)}
    \label{fig:demonstrating_model_collapse}
    \vspace{-0.3cm}
\end{figure}

\paragraph{Extended \algo{}-training on math problems.} Next, we test whether our insights from synthetic tasks also extend to real-world math problems. Specifically, we take take the same 4 base models as in the previous section, and train them on a difficult math dataset through an extended number of iterations using \algo{}. Figure~\ref{fig:demonstrating_model_collapse} demonstrates our surprising finding: while \algo{} initially increases the base model's performance at a comparable rate with ground-truth RL training, extended training using \algo{} leads to sudden performance collapse. \textbf{We observe performance collapse or degradation from extended \algo{}-training across all models and training datasets,} and record these in detail in Appendix~\ref{app:qwen2.5_math_7b_full_results} and~\ref{app:qwen3_full_results}. Given that this surprising phenomenon deserves more investigation, we study it in more detail next.

\paragraph{What happens after \algo{} peaks in performance?} 

To develop a clearer understanding of the underlying reasons for this phenomenon, in this section we investigate this \algo{}-induced model collapse closely.

\begin{figure}[t!]
    \centering
    \includegraphics[width=0.99\linewidth]{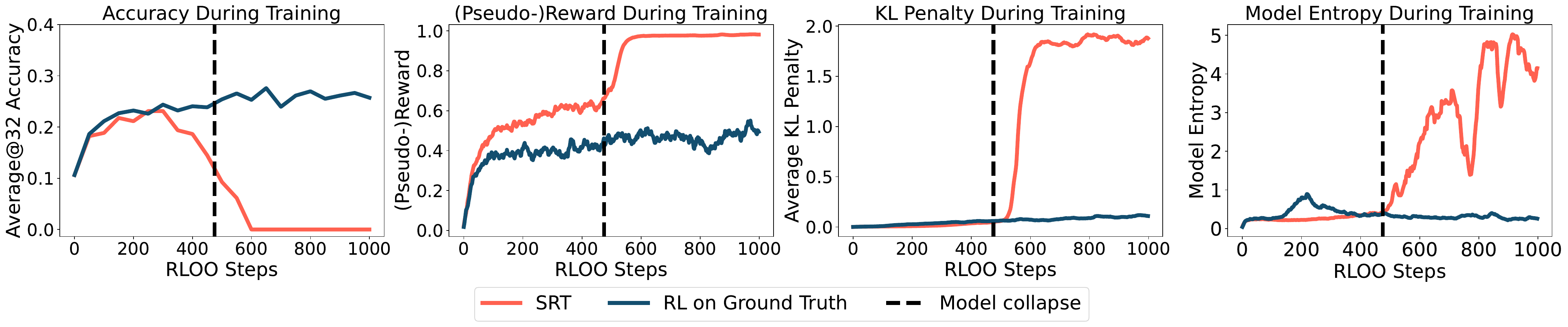}
    \caption{\footnotesize \textbf{(Self-Training Dynamics)} Extended training via \algo{} can lead to reward hacking, as demonstrated by the sudden hike in KL penalty and training (psuedo-)reward, but collapse of accuracy on the held-out test sets.}
    \vspace{-0.3cm}
    \label{fig:dynamics_of_model_collapse}
\end{figure}

We plot the training statistics of the \algo{} objective (Eqn. \ref{eqn:self-reward}) in Figure \ref{fig:dynamics_of_model_collapse}. The observed performance collapse closely coincides with a sudden increase in the \algo{} self-reward objective, implying that the optimization procedure has, in fact, maximally optimized the training objective (self-consistency majority voting), despite a decline in \textit{actual} output correctness. On the same figure, we further report the token-level average Kullback–Leibler divergence between the model under \algo{} training and the base model. We observe a sharp increase in KL divergence at the exact point when performance begins to deteriorate, indicating that the generative distribution of the model has substantially diverged from the original model.

These findings strongly suggest the occurrence of \textbf{reward hacking}---the model has learned to produce consistent responses in order to optimize its self-assigned reward, irrespective of their true correctness. Indeed, manual analysis of the model outputs (examples provided in Appnedix \ref{app:reward_hacking_examples}) confirms this hypothesis: \textbf{after collapse, the model outputs a very high entropy, essentially random, set of tokens followed by the same ``template'' final answer} (for example, \texttt{\string\boxed\{1\}}) that is nearly independent of the input prompt).
In other words, the initially strong correlation between the \algo{} objective and correctness is ultimately compromised, becoming no longer a reliable proxy signal. 
This behavior is also related to the well-known simplicity bias in neural networks~\citep{depalma2019randomdeepneuralnetworks,vallepérez2019deeplearninggeneralizesparameterfunction,mingard2020neuralnetworksprioribiased,Mingard2025Deep}, as well as the Occam's razor, where neural networks tend to find the simplest solution that generalizes to the observed signal --- in this case this leads to the same final template answer for all prompts.

\begin{figure}[h]
    \centering
    \includegraphics[width=0.99\linewidth]{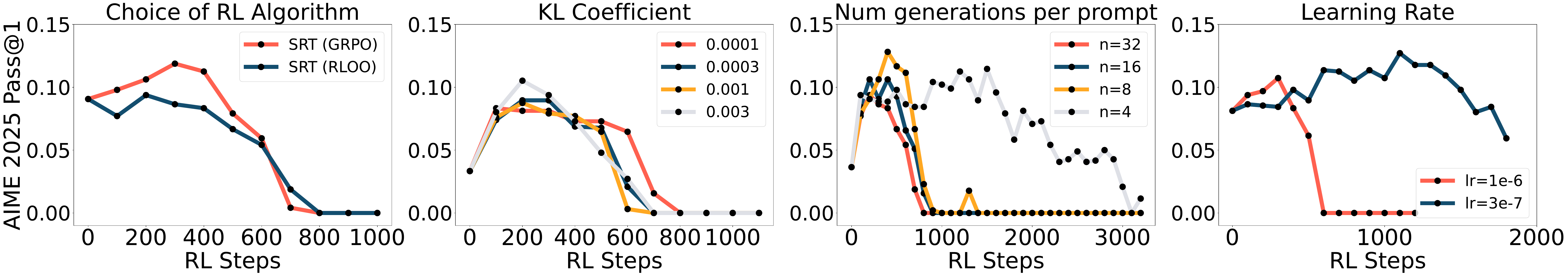}
    \caption{\footnotesize \textbf{(Sweep over training configurations)} We vary the training settings for \algo{} from the default ones described in Appendix~\ref{app:training_hparam_details} and record the observations. All experiments are run using Qwen2.5-Math-7B on DAPO. Evaluations are done using AIME 2025, but other datasets show similar behavior. In summary, choice of RL algorithm between GRPO and RLOO, and different KL coefficients do not affect model collapse significantly. Lowering learning rate slows down model collapse as expected. Finally, reducing number of generations injects noise into the majority vote, surprisingly resulting in a slow down of model collapse.}
    \label{fig:training_choice_summary}
\end{figure}

\paragraph{Additional experiments and ablations.}
We have run additional experiments and ablations to study the behavior of self-training under different training scenarios. Figure~\ref{fig:training_choice_summary} shows our main findings: \textbf{(1)} choice of RL algorithm (GRPO vs RLOO) does not affect the final outcome of \algo{}-training, \textbf{(2)} increasing KL coefficient to incentivize the model to stay close to the base policy also does not mitigate reward hacking, as the training signal from the reward hacked solution is too strong, \textbf{(3)} decreasing learning rate seems to delay model collapse but not eliminate it, and we hypothesize that prolonged training with lower learning rates would still result in complete model collapse, and \textbf{(4)} surprisingly, reducing the number of generations per prompt injects noise into the training signal, which delays quick model collapse by exploiting the majority voting answer. Additional experiments related to \algo{} training, including training curves on all test datasets and effect of tuning additional hyperparameters like entropy coefficient, can be found in Appendix~\ref{app:additional_experimental_results} and~\ref{app:detail_on_hp_ablations}. We also investigated various mitigation strategies for \algo{}-induced model collapse, including early stopping, curriculum learning, and employing a fixed teacher to generate labels -- refer to Appendix~\ref{sec5.mitigation} for our findings.

\begin{AIbox}{Takeaway 2: Self-Training benefits may not extend indefinitely}
The question of whether self-training can be extended indefinitely has mixed results: while under controllable difficulty, \algo{} can keep improving beyond the base model on progressively more difficult tasks, training on real-world math problems demonstrate the phenomenon of reward hacking --- sustained self-improvement requires developing additional regularization measures to be effective.
\end{AIbox}
\vspace{-0.1cm}

\section{Related Works}\label{sec:related_work}
\paragraph{Self Improving LLM.} Previous works \citep{zelikman2022star, wang-etal-2023-self-instruct, huang-etal-2023-large, madaan2023self,10.5555/3692070.3692326, gulcehre2023reinforcedselftrainingrestlanguage, singh2024beyond, DBLP:conf/iclr/NiIWPMRG23, DBLP:conf/emnlp/HwangKKYS24, havrilla2024glorewhenwhereimprove, pang2024language} have demonstrated the feasibility of LLMs' self-improvement over their previous iteration by training on data distilled by the previous instances of the model. Most of these approaches usually have data filtering/reranking step in the pipeline, which is often performed by the model itself \citep{wu2025metarewarding} or by training another \citep{hosseini2024vstar} verifier model.  Particularly, \citep{huang-etal-2023-large, wang2023selfconsistency,prasad2024selfconsistencypreferenceoptimization} demonstrated the feasibility of using majority voting and self-consistency to filter chain-of-thought traces that, when used as SFT training data, improve the LLM performance on downsteaming tasks. A concurrent work, \citep{zhao2025absolutezeroreinforcedselfplay}, proposes a self-evolution, self-play pipeline where an LLM generates coding problems of appropriate difficulty, solves and trains on them using RLVR. The model-generated solutions are validated by a code executor in the loop. 
Previous works have studied self-improvement by generating labels through majority voting ~\citep{prasad2024selfconsistencypreferenceoptimization, huang-etal-2023-large}, but these works are typically confined to one or a few rounds of SFT or DPO~\citep{rafailov2024directpreferenceoptimizationlanguage}. This essentially involves distilling majority voting labels from a fixed policy into the current policy over each round of training. In contrast, we explore \textit{online RLVR's} potential in self-improving LLMs where the label generating policy evolves after every gradient step. A few concurrent works~\citep{chen2025selfquestioninglanguagemodels,prabhudesai2025maximizingconfidenceimprovesreasoning,zhao2025learningreasonexternalrewards,shao2025spuriousrewardsrethinkingtraining} have also explored various forms of self-rewarding mechanisms through majority voting or a similar metric of self-consistency (e.g., token/sequence level entropy) in Online RLVR. 
Finally, in a recent work, \citep{song2025mind} formalized a generation verification gap as central to the model's ability to self-improve. Similarly, \citep{huang2025selfimprovement} proposed a ``sharpening" mechanism as the key to self-improvement. Our \algo{} pipeline builds on top of both of these intuitions. We refer the interested reader to~\citet{gao2025surveyselfevolvingagentspath} for a survey of other works associated with self-evolving LLM agents.

\paragraph{Online RLVR and Easy to Hard Generalization.} Online reinforcement learning with verifiable reward (RLVR) \citep{lambert2025tulu3pushingfrontiers} has emerged as a new paradigm of LLM post-training especially for enhancing math, coding and reasoning performances \citep{openai2024openaio1card, deepseekai2025deepseekr1incentivizingreasoningcapability, kimiteam2025kimik15scalingreinforcement, lambert2025tulu3pushingfrontiers}. Despite the success of the reasoning models, it is still unclear to what extent they can generalize beyond the difficulty of their training data distribution, a problem termed easy to hard generalization \citep{sun2024easytohard}. \citep{sun2024easytohard} shows that models can be trained to solve level 4-5 MATH\citep{hendrycks2021measuring} problems after training using a process reward model trained on MATH level 1-3 dataset. Another work \citep{lee2025selfimprovingtransformersovercomeeasytohard} explores this question and finds that transformers are capable of easy-to-hard generalization by utilizing \textit{transcendence} phenomenon \citep{zhang2024transcendence} in the context of simple addition, string copying, and maze solving using small language models. 

\paragraph{Model Collapse and Reward Hacking.} Model collapse is a well-known phenomenon in training on self-generated training data \citep{alemohammad2024selfconsuming, DBLP:journals/nature/ShumailovSZPAG24, shumailov2024curserecursiontraininggenerated, bertrand2024on, briesch2025large}, and multiple approaches related to data mixing, reliable verification, training using contrastive loss using negative samples and curriculum learning have been proposed \citep{gerstgrasser2024is, feng2025beyond, briesch2025large, song2025mind, gillman2024selfcorrectingselfconsumingloopsgenerative, setlur2024rl} to prevent models from collapsing, which previous work on LLM's easy to hard generalization \citep{lee2025selfimprovingtransformersovercomeeasytohard} also utilize. 
However, in RL paradigm, we do not directly do supervised fine-tuning on model-generated data, and it remains an open question to what extent the previous findings of model collapse apply to our RLVR setting. In our work, we show that models trained using RL on self-labeled data often suffer from actor collapsing due to reward hacking \citep{amodei2016concreteproblemsaisafety, denison2024sycophancysubterfugeinvestigatingrewardtampering} and propose a few strategies to mitigate it. Finally, a concurrent and complementary work,~\citet{zhang2025freelunchrethinkinginternal}, has also demonstrated the failure of various self-reward mechanisms to sustain self-improvement under prolonged training, which further validates the observations in this work.

\paragraph{Data Efficient RLVR.} Several concurrent works look into the data efficiency of the RLVR pipeline. Notably, \citep{wang2025reinforcementlearningreasoninglarge} shows that by just training on one example, the model can achieve performance equivalent to training on 1.2k examples from the DeepScaleR dataset \citep{deepscaler2025}. Similarly, TTRL \citep{zuo2025ttrl}, also proposes a label-free online RLVR paradigm in a test-time setting, similar to \algo{}. Our work also explores this paradigm in Section \ref{sec4.test.time.self.improvement} as an extension of SRT.

\section{Limitations and Conclusion}

In this work, we examine a simple strategy of leveraging an LLM's self-consistency to train it via an RL framework. Our experiments on synthetic reasoning tasks with controllable difficulty levels demonstrate the promise of such a framework: not only can LLMs improve their performance on these tasks, but they can also improve the quality of their self-supervision for the next training step. Our analysis, however, also reveals the limitation of leveraging self-consistency as training reward: prolonged training under such a framework can lead to reward hacking and complete model collapse. Future investigations could explore how to develop more robust forms of verification, extend self-verification to other domains like coding, and curriculum learning strategies that expose the model to problems of only the right difficulty. Additionally, leveraging LLM-as-judges or generative verifiers to improve the training signal, or employing additional consistency regularization between the chain-of-thought and the final answer to mitigate reward hacking can be promising future research. We leave these and other promising directions for sustained self-improvement for future work.

\section*{Acknowledgements}
This work has greatly benefited from the use of Delta's advanced computing and data resource supported by the National Science Foundation (OAC 2005572) and the State of Illinois, as part of ACCESS-approved compute grants~\citep{access_compute}. We also appreciate the computing resources of Bridges-2~\citep{psc_computing} at Pittsburgh Supercomputing Center through ACCESS allocation CIS240901 from the Advanced Cyberinfrastructure Coordination Ecosystem: Services \& Support (ACCESS) program, which is supported by National Science Foundation grants \#2138259, \#2138286, \#2138307, \#2137603, and \#2138296. Overall, this project used ACCESS grants CIS240901, CIS230366, CIS250216, and CIS250428 for its compute resources. Moreover, FT was partially supported by the U.S. Army Futures Command under Contract No. W519TC-23-C-0030 during the project. Andrea Zanette was partially supported by the National Science Foundation under Grants CCF-2106778 and DMS-2134080.

The authors thank Brandon Pusateri, Jillian Lehosky, and Greg Bauer from ACCESS Support Staff for their incredible help at approving supplements and renewals for ACCESS compute grants throughout this project. Moreover, the work would not have finished so quickly without the help of Brett Bode from NCSA Delta Support Staff, who provided the authors with critical help in properly utilizing the Delta cluster.  FT gratefully acknowledges Daman Arora, Yuda Song, Yiding Jiang, Ruiqi Zhang, Zhaoyi Zhou, Guanning Zeng, Yutong He and other members of the Zanette, Russ, Auton, and AIRe lab for feedback and suggestions received on earlier versions of this work.

\bibliography{bibliography}
\bibliographystyle{iclr2026_conference}

\newpage
\newpage
\appendix

\section{Definition of Generation-Verification Gap}
\label{app:gen-ver-gap}
The single-generation accuracy is defined as:
\[
\genacc = \mathbb{E}_{\prompt\sim\promptDist, \response\sim\policy(\cdot|\prompt)}[\mathbf{1}(\response=\trueanswer)],
\]
where $\trueanswer$ is the correct solution.
A verifier function $\verifier$ selects one candidate from multiple generations:
\[
\verifier\left(\prompt, \{\response^{(1)},\dots,\response^{(n)}\}\right)\in\{\response^{(1)},\dots,\response^{(n)}\}.
\]
We define verification accuracy as:
\[
\veracc = \mathbb{E}_{\prompt\sim\promptDist}\left[\mathbf{1}\left(\verifier\left(\prompt,\{\response^{(1)},\dots,\response^{(n)}\}\right)=\trueanswer\right)\right].
\]
We say that a positive \emph{generation-verification gap} occurs whenever $
\veracc > \genacc.
$
Such a gap indicates the verifier's greater proficiency in recognizing correct solutions within a set of candidates compared to the generator independently generating correct answers.

\subsection{Generation Verification Gap Through Majority Voting}
\begin{figure}[h!]
    \centering
    \includegraphics[width=0.95\linewidth]{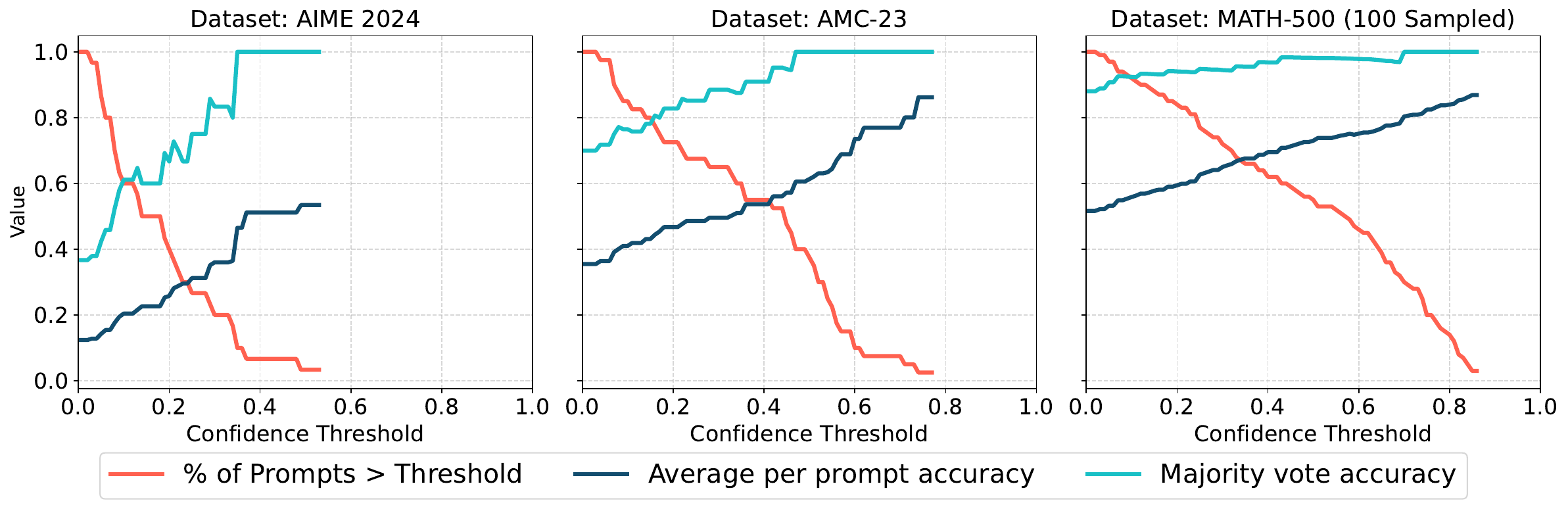}
    \caption{Three of our test datasets display evidence of generation verification gap through majority voting. \textbf{The positive gap between majority voting accuracy and per prompt accuracy means LLMs can utilize this gap as a learning signal to improve their average accuracy}. x axis refers to the threshold cut off for self-consistency (proportion of answers that are majority voted answers). Higher x value refers to more self-consistent LLM outputs. For example, at \texttt{x = 0.4}, we only keep LLM responses where at least 40\% of the (properly parsed) responses were the majority voted answer. On y axis, \textbf{average per prompt accuracy} refers to the average number of correct answers out of the (successfully parsed) responses, across 32 generations per question (computed on a per prompt basis and then averaged over the dataset). \textbf{Majority vote accuracy} refers to how often the majority vote is the correct answer for that prompt (then averaged over the dataset). \textbf{\% of Prompts} was computed over all 32 generations (\textit{not} on the parsable answers for a fair comparison). If there is no prompt left over a certain threshold, the line plot ends there. }
    \label{fig:generation_verification_gap}
\end{figure}

\newpage

\section{Details on Implementation and Training}

\subsection{RL Algorithms} 

We use two RL algorithms in this work: RLOO\citep{ahmadian-etal-2024-back, Kool2019Buy4R} and GRPO~\citep{shao2024deepseekmathpushinglimitsmathematical}. Recent work~\citep{Oertell2024Heuristics} has shown that heuristic policy gradient algorithms like GRPO can produce unexpected results by increasing or decreasing reasoning performance even under random rewards, where policy gradient should be zero in expectation, and that RLOO does not have this problem. Since \algo{} is compatible with both RL algorithms, we experiment with both and observe no noticeable difference in the resulting behavior.

In our implementation (\texttt{verl}), we use the following RL objective for both RLOO and GRPO:

\begin{align*}
\mathcal{J}_{\text{GRPO}}(\theta) 
= \mathbb{E}_{x \sim \mathcal{D},\ \{y_i\}_{i=1}^G \sim \pi_{\theta_{\text{old}}}(\cdot|x)} 
\Bigg[ 
    \frac{1}{G} \sum_{i=1}^G \frac{1}{|y_i|} \sum_{t=1}^{|y_i|} 
    \min \Big( & w_{i,t}(\theta) \hat{A}_{i,t}, \\
    & \text{clip}(w_{i,t}(\theta),\ 1 - \varepsilon,\ 1 + \varepsilon) \hat{A}_{i,t} 
    \Big) 
\Bigg]
\end{align*}

where $w_{i, t}(\theta)$ is the importance ratio, defined as:

$$w_{i,t}(\theta) = \frac{
    \pi_\theta(y_{i,t} \mid x,\ y_{i,<t})
}{
    \pi_{\theta_{\text{old}}}(y_{i,t} \mid x,\ y_{i,<t})
}$$

Since we operate fully on-policy, i.e., one RL step per one batch of generated rollouts, this is always one in our experiments. The same advantage defined at a sequence level is applied to each token in the sequence, so henceforth we will drop the $t$ from the notation as well.

The main difference between GRPO and RLOO then stems from their use of different advantage functions. RLOO objective uses the following advantage function:

$$\frac{1}{G}\sum_{i=1}^{G}[R(y_{(i)},x) - \frac{1}{G-1}\sum_{j\neq k}R(y_{(j)},x)]$$

whereas GRPO uses the following advantage function:

$$\hat{A}_i = \frac{
    r(x, y_i) - \text{mean}\left( \{ r(x, y_i) \}_{i=1}^G \right)
}{
    \text{std}\left( \{ r(x, y_i) \}_{i=1}^G \right)
}$$

Here $G$ is the number of online samples generated. Both RLOO and GRPO create a dynamic baseline for each sample without needing a separate value function, effectively estimating the expected return on-the-fly during training. Not using a separately trained value network (as opposed to PPO) makes the training much simpler for both algorithms.

In our implementation, we did not add the KL penalty to the loss function, but rather to the reward itself while running RLOO, following recent work such as~\citet{tang2025pitfallskldivergencegradient}. In verl framework, this can be configured using \texttt{algorithm.use\_kl\_in\_reward=True} and \texttt{actor\_rollout\_ref.actor.use\_kl\_loss=False}. However, this does not work for GRPO due to advantage normalization by the standard deviation, and so for GRPO we add KL penalty to the loss function directly. To estimate KL penalty, we use the low variance KL estimator proposed by~\citet{schulman2020klapprox}:

$$\mathbb{D}_{\text{KL}}(\pi_\theta || \pi_\text{ref}) \approx \frac{\pi_\text{ref} (y | x)}{\pi_\theta (y|x)} - 1 - \log \left( \frac{\pi_\text{ref} (y | x)}{\pi_\theta (y|x)}\right)$$

\paragraph{Sampling.} For all experiments, we kept the generation \texttt{temperature} to 1.0, \texttt{top\_k} to -1, and \texttt{top\_p} to 1 for rollouts generated during RL rollouts. Decoding temperature used for validation varies in different settings, see Appendix~\ref{app:training_hparam_details} and~\ref{app:sampling_temp} for more discussion. We cut off maximum prompt length at 1024 and maximum response length to 3072 (note: Qwen2.5-Math-7B models support a maximum context window of 4096). 

\subsection{GPU Infrastructure}
All experiments in this work were conducted using either a single node consisting of 8 NVIDIA H200 GPUs (141 GB of GPU memory per GPU) or a single node consisting of 4 NVIDIA GH200 GPUs (96 GB of GPU memory per GPU). All experiments can be replicated in single-node training, and we did not, in fact, utilize multinode training. In total, this work consumed $\sim$15000 GPU hours (including preliminary studies and failed runs). All the final results listed in this paper can be replicated within 2000 H200 GPU hours.

\subsection{Details on Training Settings}
\label{app:training_hparam_details}

We choose Family Relationships, Bitwise Arithmetic, and Knights \& Knaves \citep{xie2024memorization} tasks from Reasoning Gym~\citep{stojanovski2025reasoninggymreasoningenvironments} for our experiments. Examples for each task is shown in Appendix \ref{app:reasoning_gym_example_tasks}. 

For the \textbf{Bitwise Arithmetic} task, \emph{Level} refers to the \texttt{difficulty} parameter. The model was first trained with level~2 data for 950 steps, reaching 97\% accuracy. We then used this initialized model to train with \algo{} on levels~3 and~4. 

For the \textbf{Family Relationships} task, we trained a model to 99\% accuracy on the level~4 dataset. Here, \emph{Level} corresponds to the parameters \texttt{min\_family\_size} and \texttt{max\_family\_size}, both set to~4. We then applied \algo{} on level 5. 

For the \textbf{Knights \& Knaves} task, we varied only the \texttt{n\_people} parameter as the difficulty control. We first trained a model with difficulty level~2 to 99\% accuracy, and then used that checkpoint to further experiment with \algo{}, climbing from level 2 to 3, 3 to 5, 5 to 7 and 7 to 9. \textbf{We only report levels 7 and 9 in the paper since they are the highest level difficulty among our experiments}. 

Across all multi-level experiments, we applied \algo{} progressively. For example, in Bitwise Arithmetic we trained on level~2 with ground truth supervision; then, starting from the level~2 checkpoint, we applied \algo{} on level~3; finally, we repeated \algo{} again on level~4 using the checkpoint from level~3. For comparing against \textbf{\algo{} with fixed teacher}, we use the same starting policy (Qwen3-4B-Base trained with ground-truth on the easiest difficulty level on each task) and generate the same number of rollouts per prompt using temperature 1.0 and perform majority voting among these rollouts to generate our pseudo-labels.

\paragraph{Default training hyperparameters for Reasoning Gym tasks.} For all Reasoning Gym experiments, we used the \texttt{Qwen3-4B-Base} model, with GRPO as the main algorithm. The learning rate was set to \texttt{1e-6} and the KL penalty to \texttt{0.0001}. For all experiments, we used 32 rollouts per prompt for training and 16 rollouts for evaluation.

\paragraph{Default training hyperparameters for Math Datasets.} 

\begin{itemize}
    \item \textbf{Qwen2.5-Math-7B}: Learning rate $10^{-6}$, KL penalty coefficient $0.001$, decoding temperature for training and evaluation rollouts $1.0$, top-p $1.0$, and no top-k sampling.
    \item \textbf{Qwen3-14B-Base}: Same default hyperparameter setting as Qwen2.5-Math-7B.
    \item \textbf{Llama-3.1-8B-Instruct}: Learning rate $10^{-7}$, KL penalty coefficient $0.001$, decoding temperature for training is $1.0$ and evaluation is $0.0$, top-p $1.0$, and no top-k sampling. We subsample the Big-Math-RL-Verified dataset to only keep prompts that has average Llama-3.1-8B-Instruct pass rate between 0.3 and 0.7, since the model is unable to improve during training otherwise.
    \item \textbf{Deepseek-Math-7B-Instruct}: Learning rate $10^{-7}$, KL penalty coefficient $0.001$, decoding temperature for training is $1.0$ and evaluation is $0.7$ (following official protocol), top-p $1.0$, and no top-k sampling. 
\end{itemize}

\subsection{Code and Dataset Release}

For the ease of reproducing the results in this paper, we release the code and datasets used in this work. Specifically, we release four datasets (the others being publicly available):

\begin{enumerate}
    \item Deduplicated version of the DAPO dataset, containing only unique prompts: \url{https://huggingface.co/datasets/ftajwar/deduplicated_dapo_dataset}

    \item The test datasets for math tasks compiled together: \url{https://huggingface.co/datasets/ftajwar/srt_test_dataset}

    \item Individual training and evaluation datasets for various levels of the reasoning gym tasks (bitwise arithmetic, knights and knaves, family relationships) can be found here: \url{https://huggingface.co/collections/ftajwar/self-rewarding-llm-training-6835218091832c3664176553}.

    \item The easy subset of DAPO, chosen based on per prompt pass rate calculated using 32 responses for each prompt from a Qwen2.5-Math-7B model (at temperature 1.0 and no top\_p or top\_k sampling): \url{ftajwar/dapo_easy_one_third_sorted_by_pass_rate}

    \item The easy subset of DAPO, chosen based on per prompt frequency of the majority answer from a Qwen2.5-Math-7B model using the same generation setting as above: \url{https://huggingface.co/datasets/ftajwar/dapo_easy_one_third_sorted_by_pass_rate}
\end{enumerate}

Finally, our code is available at: \url{https://github.com/tajwarfahim/srt}. For a summary of project related assets, visit our project website at \url{https://self-rewarding-llm-training.github.io/}.

\newpage

\section{Additional Experimental Results}
\label{app:additional_experimental_results}

\subsection{Qwen2.5-Math-7B}
\label{app:qwen2.5_math_7b_full_results}

\begin{figure}[h!]
    \centering
    \includegraphics[width=0.99\linewidth]{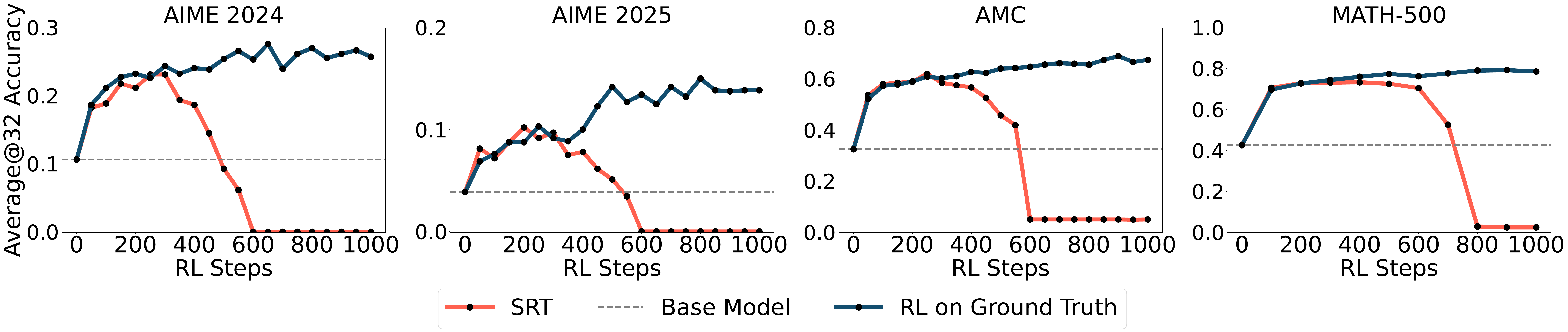}
    \caption{\footnotesize \textbf{(Individual test set performance during training on DAPO)} We record the average@32 accuracy during training Qwen2.5-Math-7B on DAPO, on three heldout test sets: AIME 2024, AIME 2025 and AMC. In all three cases, \algo{} performance collapses, while training with ground truth keeps improving steadily.}
    \label{fig:dapo_detailed_results}
\end{figure}

\begin{figure}[h!]
    \centering
    \includegraphics[width=0.99\linewidth]{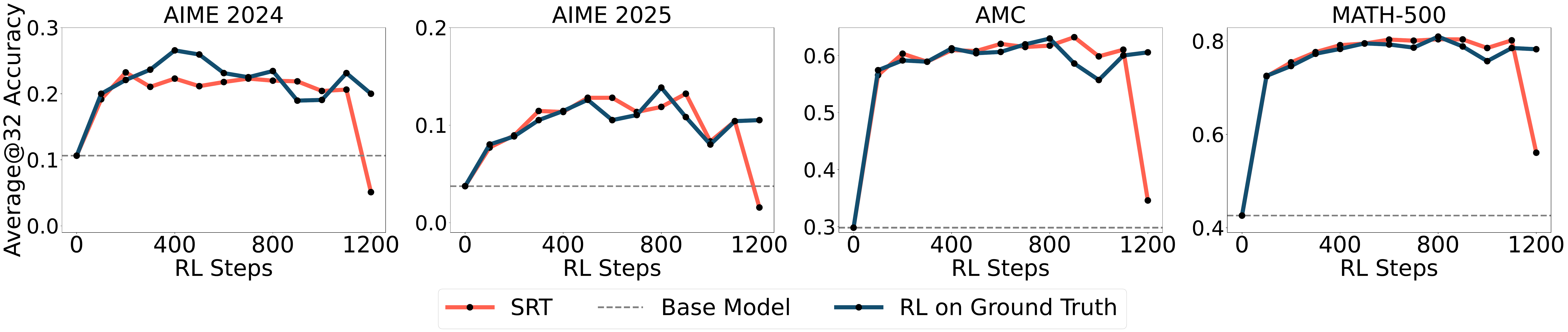}
    \caption{\footnotesize \textbf{(Individual test set performance during training on MATH-12K)} We record the average@32 accuracy during training Qwen2.5-Math-7B on MATH-12K, on three heldout test sets: AIME 2024, AIME 2025 and AMC. We also evaluate intermediate checkpoints on MATH-500 since we are training on MATH-12K (we could not do this for other training datasets due to a lack of computational resources). In all 4 heldout test sets, \algo{} results in similar performance gain as one would obtain from training with ground truth labels. However, performance collapses after 1200 RL steps, similar to our other observations.}
    \label{fig:math_detailed_results}
\end{figure}

\begin{figure}[h!]
    \centering
    \includegraphics[width=0.99\linewidth]{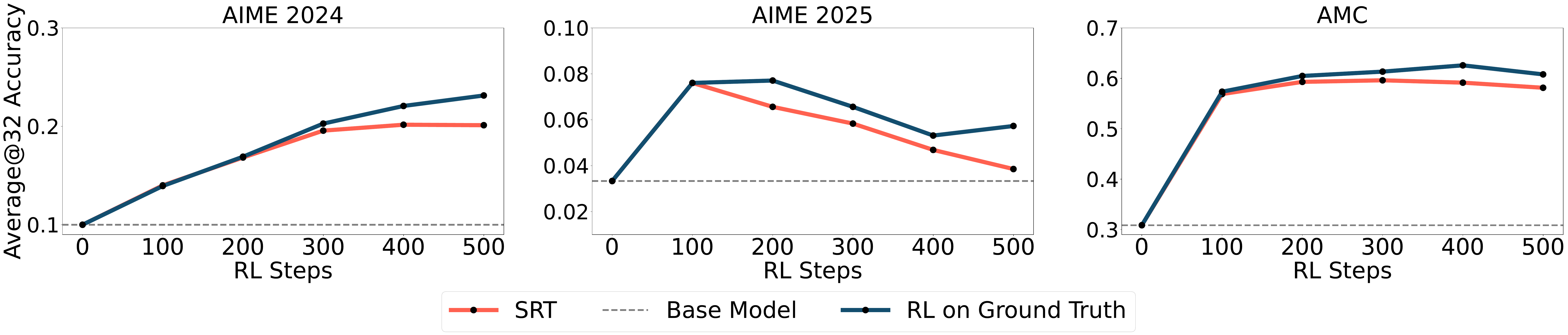}
    \caption{\footnotesize \textbf{(Individual test set performance during training on AIME (1983-2023))} We record the average@32 accuracy during training Qwen2.5-Math-7B on AIME (1983-2023), on three heldout test sets: AIME 2024, AIME 2025 and AMC. \algo{} performs similarly or better compared to training with ground truth labels over 10 epochs of training.}
    \label{fig:aime_detailed_results}
\end{figure}

Here we compare the performance of training Qwen2.5-Math-7B with \algo{} and RL with ground truth on each individual test set for the sake of completeness. Figures \ref{fig:dapo_detailed_results}, \ref{fig:math_detailed_results}, and \ref{fig:aime_detailed_results} show the detailed results when we train on DAPO, MATH-12K, and AIME (1983-2023) respectively. Additionally, when training on MATH-12K and DAPO, we also evaluate the intermediate checkpoints on the heldout set MATH-500, which is reported in Figure \ref{fig:math_detailed_results}. \textbf{Since MATH-500 contains 500 examples, calculating average@32 accuracy becomes expensive, and hence we could not use it as a test set for all our training setups}.

\pagebreak 

\subsection{Qwen3-14B-Base}
\label{app:qwen3_full_results}

In addition to Qwen2.5-Math-7B, we apply our algorithm on another LLM --- namely Qwen3-14B-Base~\citep{yang2025qwen3technicalreport}. We choose the base model since it has not gone through additional post-training on reasoning tasks, unlike the Qwen3-14B model. Additionally, this is a significantly larger model with a different pre-training, making it suitable for testing our algorithm's effectiveness.

\begin{figure}[h!]
    \centering
    \includegraphics[width=0.99\linewidth]{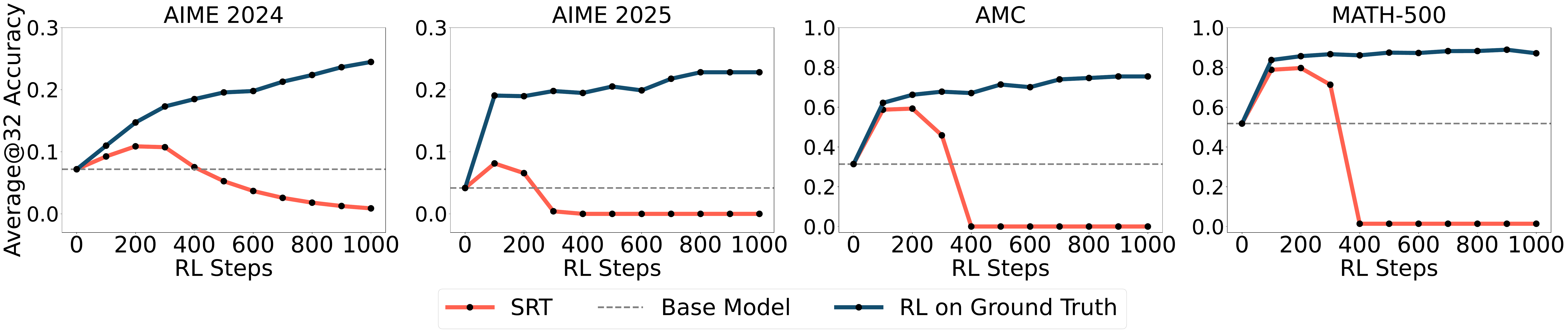}
    \caption{\footnotesize \textbf{(Individual test set performance during Qwen3-14B-Base on DAPO)} We record the average@32 accuracy during training a Qwen3-14B-Base model on DAPO, on four heldout test sets: AIME 2024, AIME 2025, AMC, and MATH-500.}
    \label{fig:dapo_detailed_results_qwen3}
\end{figure}

\begin{figure}[h!]
    \centering
    \includegraphics[width=0.99\linewidth]{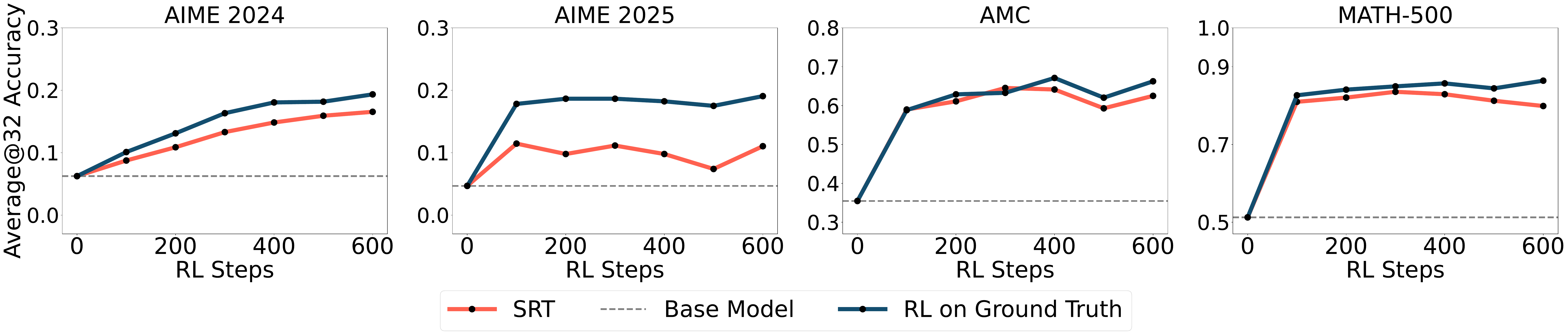}
    \caption{\footnotesize \textbf{(Individual test set performance during training Qwen3-14B-Base on MATH-12K)} We record the average@32 accuracy during training a Qwen3-14B-Base model on MATH-12K, on four heldout test sets: AIME 2024, AIME 2025, AMC, and MATH-500.}
    \label{fig:math_12k_detailed_results_qwen3}
\end{figure}

\begin{figure}[h!]
    \centering
    \includegraphics[width=0.99\linewidth]{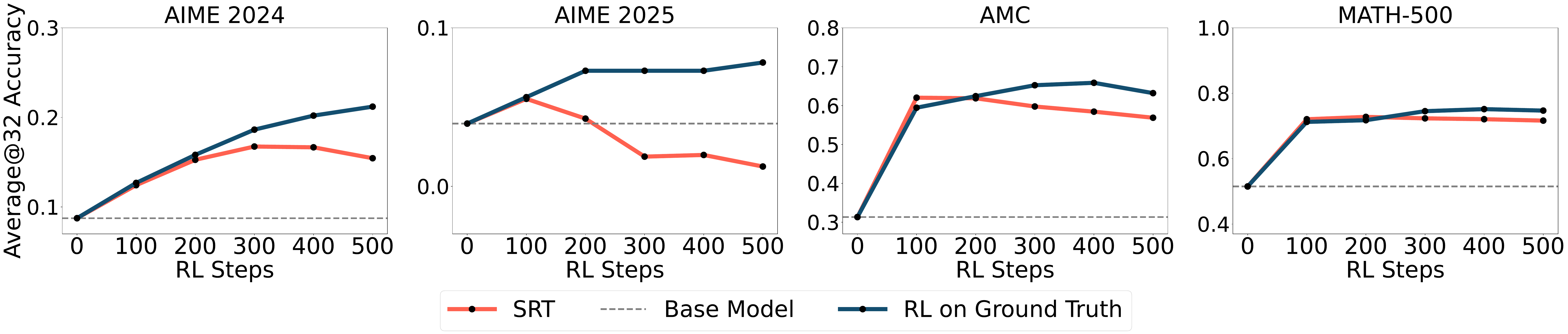}
    \caption{\footnotesize \textbf{(Individual test set performance during training Qwen3-14B-Base Model on AIME (1983-2023)} We record the average@32 accuracy during training a Qwen3-14B-Base model on AIME (1983-2023), on four heldout test sets: AIME 2024, AIME 2025, AMC, and MATH-500.}
    \label{fig:aime_detailed_results_qwen3}
\end{figure}

Figures \ref{fig:dapo_detailed_results_qwen3}, \ref{fig:math_12k_detailed_results_qwen3}, and \ref{fig:aime_detailed_results_qwen3} shows our results with DAPO, MATH-12K, and AIME (1983-2023) used as training dataset respectively. Our experiments with Qwen3-14B-Base mostly follows similar patterns as Qwen2.5-Math-7B: \algo{} maintains stable performance on MATH-12K, mixed results on AIME (1983-2023), and performance collapse on DAPO.

\pagebreak 

\section{Effect of Decoding Temperature (Qwen2.5-Math-7B)}
\label{app:sampling_temp}

\begin{figure}[h!]
  \centering
  \begin{subfigure}[]{0.48\textwidth}
    \centering
    \includegraphics[width=\textwidth]{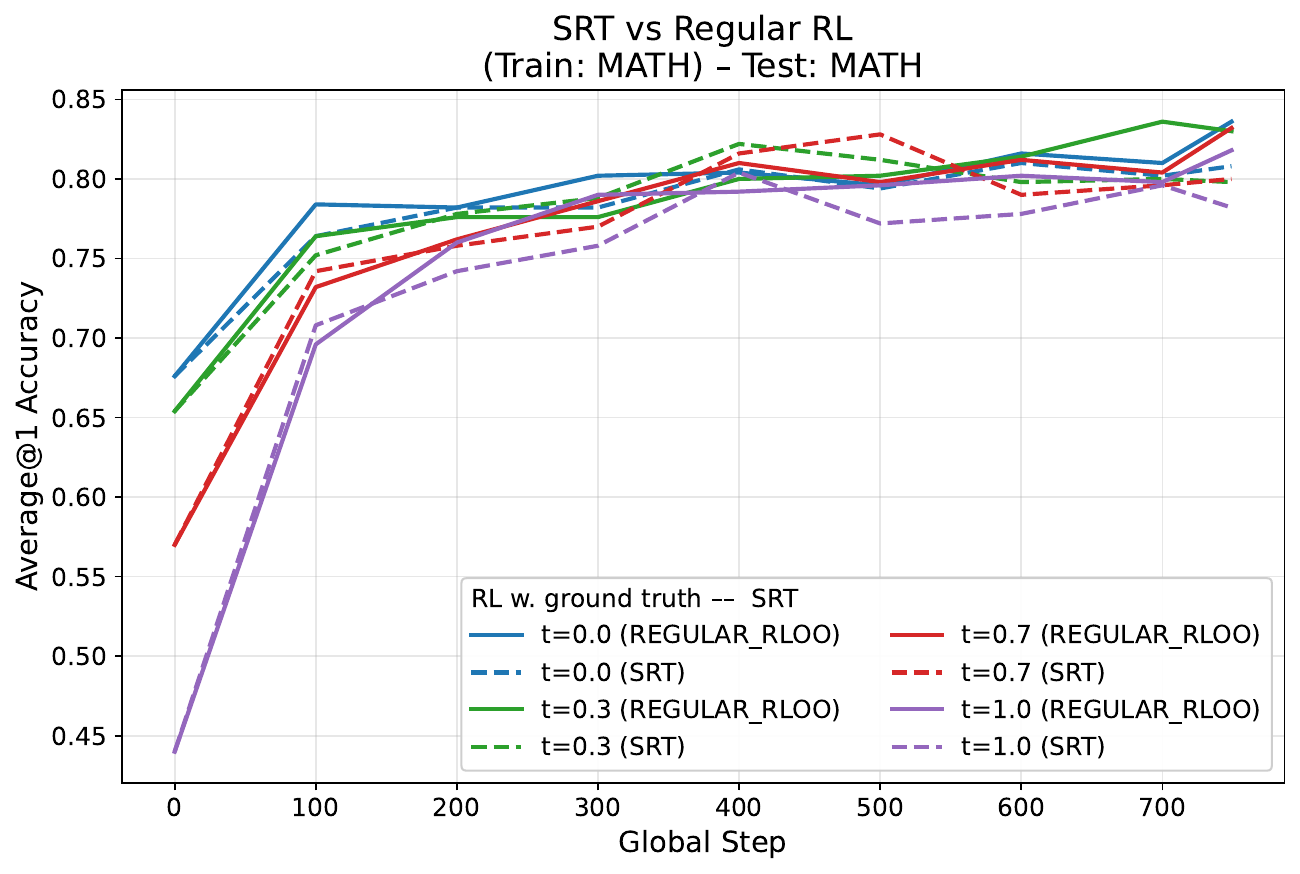}
    \caption{Math-500 evaluation accuracy during training, when decoded under various temperature (Qwen2.5-Math-7B).}
    \label{fig:tp-1}
  \end{subfigure}
  \hfill
  \begin{subfigure}[]{0.48\textwidth}
    \centering
    \includegraphics[width=\textwidth]{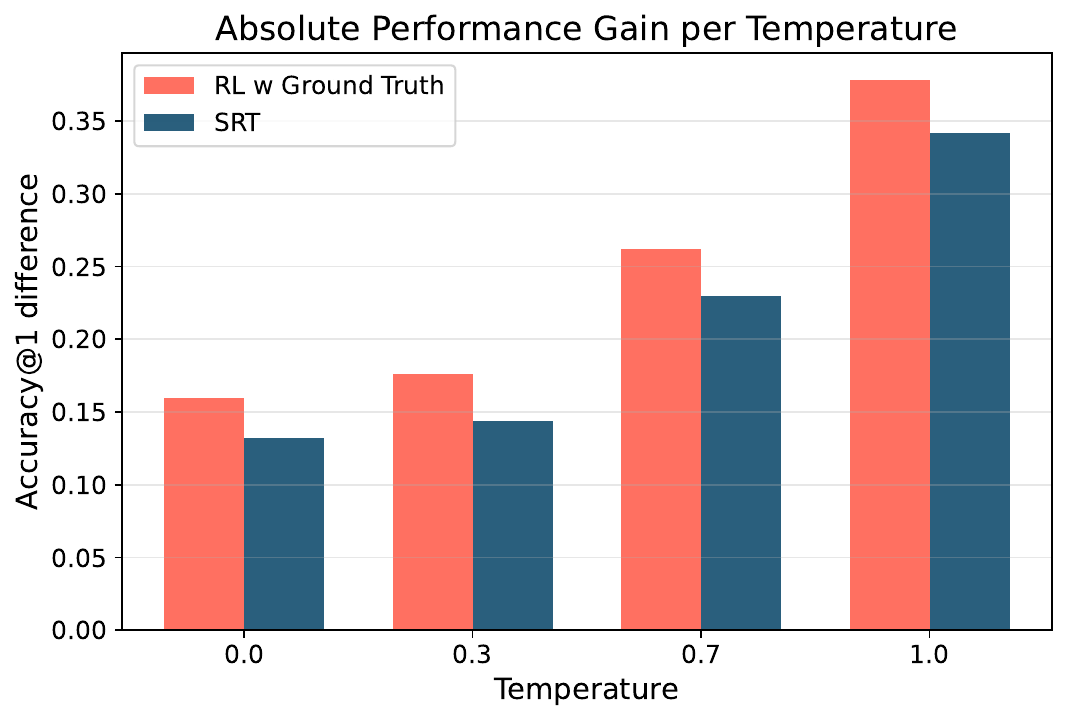}
    \caption{Absolute performance improvement after one epoch of training when decoded under various temperatures (Qwen2.5-Math-7B).}
    \label{fig:tp-2}
  \end{subfigure}
  \caption{Our method (SRT) performs consistently regardless of decoding temperature for validation (Figure \ref{fig:tp-1}). All experiments are run using Qwen2.5-Math-7B as the base model, trained on MATH-12K and tested on MATH-500. Notice that even though the performance is low initially at high temperature, at the later stages, they plateau around the same point. Figure \ref{fig:tp-2} shows the absolute gain when decoded under different temperatures. Note that decoding with higher temperature might give the impression of a larger gain compared to low-temperature decoding. However, the evaluation curves during training resulting from \algo{} and RL with ground-truth look almost identical regardless of decoding temperature, which is one of our main observations in this work.}
  \label{fig:t-p-ablation}
\end{figure}

\section{Additional Self-Training Metrics} \label{app:parsability_metrics}

\begin{figure}[h]
    \centering
    \includegraphics[width=0.99\linewidth]{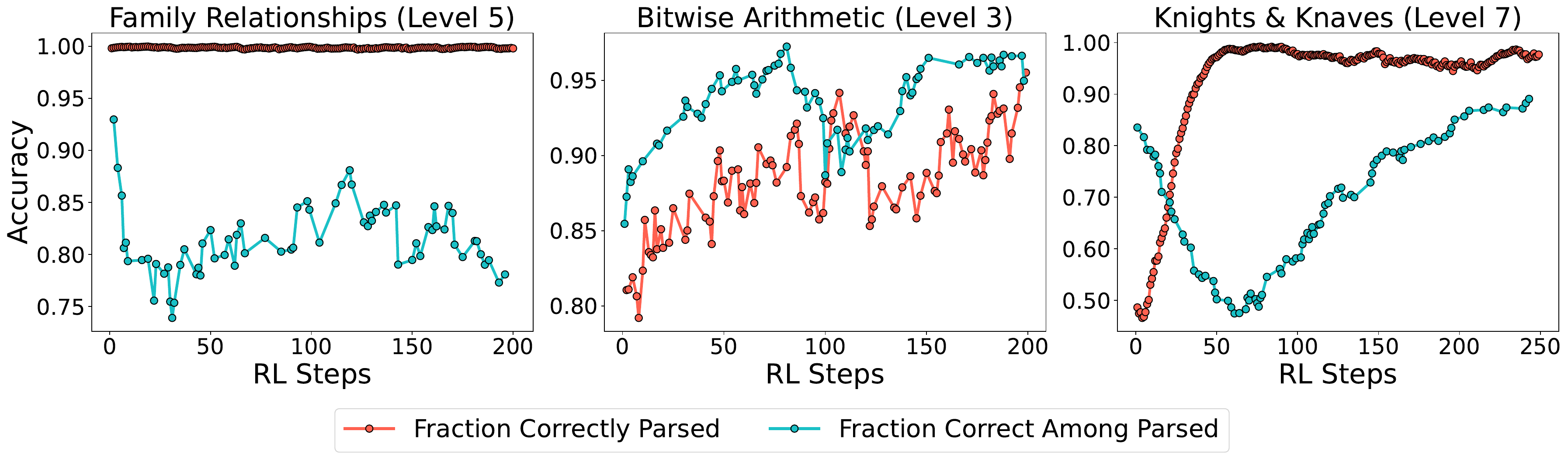}
    \caption{\footnotesize \textbf{(Tracking format following success rate during \algo{}-training on Reasoning Gym)} In order to track whether \algo{} is teaching reasoning strategies beyond just formatting the final answer correctly, we track two additional metrics throughout \algo{}-training: fraction of generations among all generations where the final answer is parseable and fraction of generations among those that are parseable where the final answer is correct. We see that due to training with RLVR on an easier level of difficulty, the starting policy can already format most generations correctly, and in the case of Knights \& Knaves, fraction of correct responses keeps increasing even after fraction of properly formatted (and thus parseable) responses have saturated. This shows that the model learns reasoning strategies beyond formatting rules.}
    \label{fig:self_labeling_metrics}
\end{figure}

We are interested to know if \algo{} teaches actual reasoning strategies beyond just proper formatting rules necessary for extracting the final answer. To do so, we track two additional metrics throughout \algo{}-training: fraction of generations among all generations where the final answer is parseable and fraction of generations among those that are parseable where the final answer is correct. Figure~\ref{fig:self_labeling_metrics} summarizes our findings on Reasoning Gym: We see that due to training with RLVR on an easier level of difficulty, the starting policy can already format most generations correctly, and in the case of Knights \& Knaves, fraction of correct responses keeps increasing even after fraction of properly formatted (and thus parseable) responses have saturated. 

\pagebreak

\section{More Details on Baselines} \label{app:baseline_details}

\paragraph{Baseline Implementation. }
For all three methods (SFT, DPO~\citep{rafailov2024directpreferenceoptimizationlanguage}, ScPO~\citep{prasad2024selfconsistencypreferenceoptimization}), we sweep over three learning rates ($10^{-5}$, $10^{-6}$ and $10^{-7}$) and pick the checkpoint with the highest validation score. The best checkpoint with highest validation in the SFT stage has been used to initialize the DPO/ScPO training. We also train DPO with the above mentioned learning rates and picked the best score. For DPO and ScPO \citep{prasad2024selfconsistencypreferenceoptimization}, we used $\beta = 0.1$, which we also found through sweep over (0.1, 0.3 and 0.5). Moreover, we add a negative log-likelihood loss with weight $1.0$ to the DPO and ScPO losses to stabilize them, similar to RPO~\citep{pang2024iterativereasoningpreferenceoptimization}. We do not train for more than 1 epoch to prevent overfitting/unintentional unalignment~\citep{tajwar2024preferencefinetuningllmsleverage,razin2025unintentionalunalignmentlikelihooddisplacement} and fair comparison with \algo{}. ScPO was also performed for one round.

\begin{table}[h!]
\centering
\renewcommand{\arraystretch}{1.2}
\begin{tabular}{@{}llcc@{}}
\toprule
\textbf{Train Dataset} & \textbf{Method} & \textbf{AMC/AIME} & \textbf{MATH500} \\
\midrule
\multirow{5}{*}{\textbf{MATH}} 
    & SFT             & 0.18 & 0.75 \\
    & ScPO            & 0.20 & 0.72 \\
    & DPO             & 0.23 & 0.74 \\
    & \textbf{SRT (Ours)} & \textbf{0.32} & \textbf{0.80} \\
\midrule
\multirow{5}{*}{\textbf{DAPO}} 
    & SFT             & 0.18 & 0.75 \\
    & ScPO            & 0.20 & 0.72 \\
    & DPO             & 0.21 & 0.76 \\
    & \textbf{SRT (Ours)} & \textbf{0.31} & 0.75 \\
\midrule
\multirow{2}{*}{\textbf{Base Model}} 
    & Accuracy     & 0.15 & 0.42 \\
    & Majority@32 Acc & 0.20 & 0.79 \\
\bottomrule
\end{tabular}
\vspace{4pt}

\caption{Comparison of different methods trained on either the MATH or DAPO dataset. Performance is evaluated on AMC/AIME24, 25 (average accuracy@32) and MATH500 (average accuracy@1). Notice that majority@32 accuracy scores are not directly comparable with the other accuracy metrics listed in the table.}
\label{tab:app_baseline_table}
\end{table}

\paragraph{Dataset Curation} For DPO and ScPO we labeled the most consistent response as the positive example and the least consistent response as the negative example for each question. Moreover, we only kept the instances where $w(x)$, the variance based weighing parameter \citep{prasad2024selfconsistencypreferenceoptimization}, was greater than 2. 

\newpage

\section{Detailed Experiment Results on Different Training Settings}
\label{app:detail_on_hp_ablations}

\subsection{GRPO vs RLOO}

\begin{figure}[h!]
    \centering
    \includegraphics[width=0.98\linewidth]{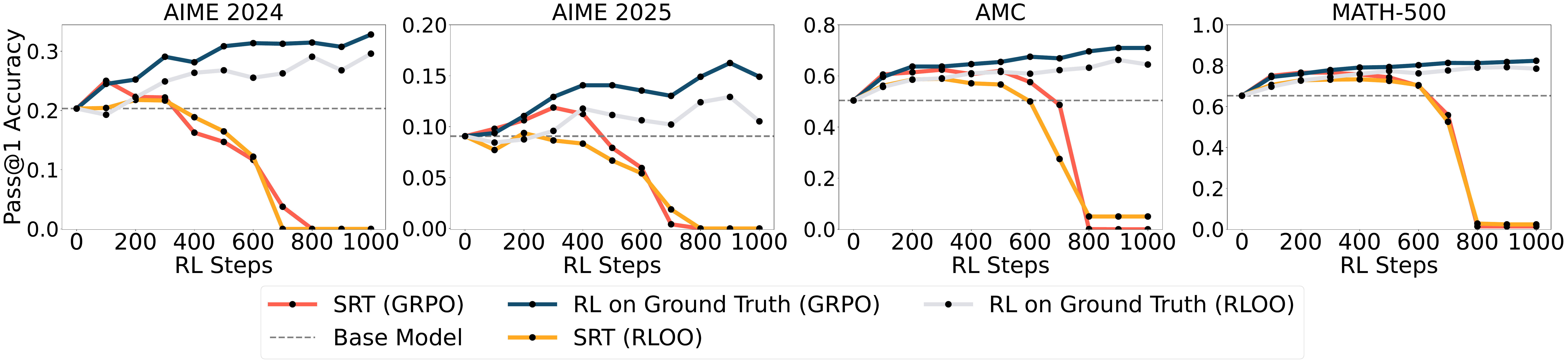}
    \caption{\footnotesize \textbf{(GRPO vs RLOO comparison)} We compare the behavior of \algo{} under two different RL optimization algorithm: GRPO vs RLOO. All experiments use a Qwen2.5-Math-7B model trained on DAPO, with the other hyperparameters being the default ones described in Appendix~\ref{app:training_hparam_details}. While \algo{} with GRPO seems to achieve higher performance than that of \algo{} employing RLOO, ultimately prolonged training using both algorithms lead to reward hacking and model collapse on all test datasets.}
    \label{fig:grpo_vs_rloo_all_test_datasets}
\end{figure}

Figure~\ref{fig:grpo_vs_rloo_all_test_datasets} shows our experiment comparing how \algo{} behaves with different RL algorithms. In particular, we test two algorithms: GRPO and RLOO. While GRPO seems to achieve higher performance, both GRPO and RLOO training with \algo{}-reward leads to model collapse at similar number of steps --- showing that the choice of the RL algo does not influence model collapse.

\subsection{Different KL Penalty Coefficients}

\begin{figure}[h!]
    \centering
    \includegraphics[width=0.98\linewidth]{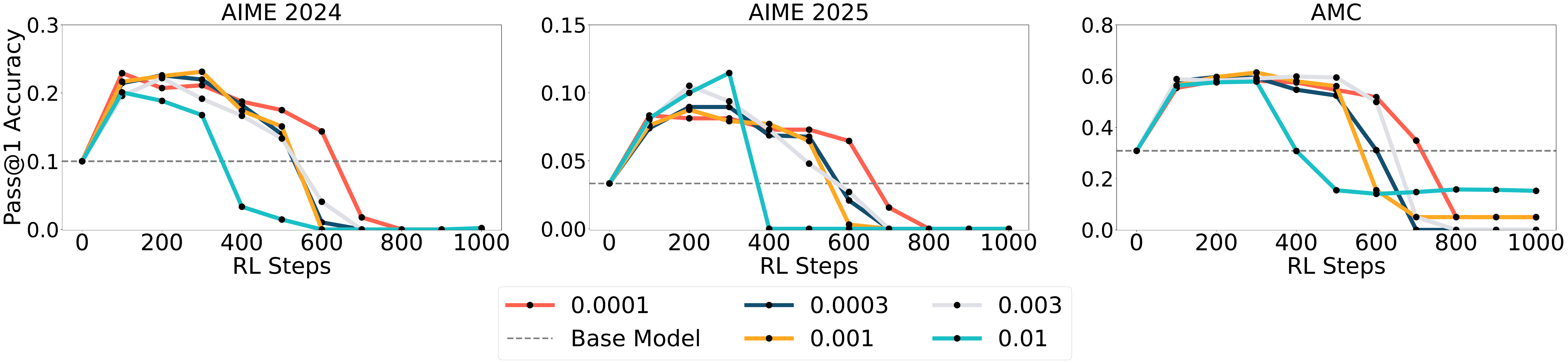}
    \caption{\footnotesize \textbf{(\algo{} with different KL penalty coefficients)} We compare the behavior of \algo{} with different KL penalty coefficients and report performance on all test datasets. All experiments here use a Qwen2.5-Math-7B model trained with RLOO on the DAPO dataset, with the hyperparameters other than KL penalty coefficient being the default ones described in Appendix~\ref{app:training_hparam_details}. Stronger KL penalty does not prevent or delay model collapse.}
    \label{fig:kl_coefficient_all_datasets}
\end{figure}

The most straightforward way of preventing reward hacking is to add a strong KL penalty to the training objective. In Figure~\ref{fig:kl_coefficient_all_datasets}, we explore this idea: to our surprise, we don't find a higher KL penalty coefficient to delay or prevent model collapse. We attribute this to the reward hacking training signal being too strong to be overcome by the KL regularization.

\newpage 

\subsection{Learning Rate}

\begin{figure}[h!]
    \centering
    \includegraphics[width=0.98\linewidth]{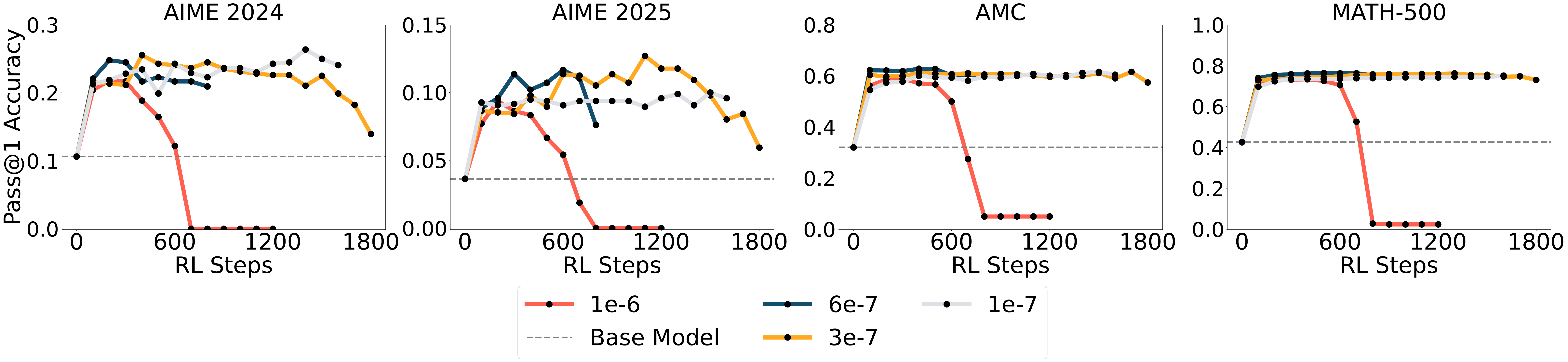}
    \caption{\footnotesize \textbf{(\algo{} with different learning rate)} We compare the behavior of \algo{} with different learning rate and report performance on all test datasets. All experiments here use a Qwen2.5-Math-7B model trained with RLOO on the DAPO dataset, with the hyperparameters other than learning rate being the default ones described in Appendix~\ref{app:training_hparam_details}. Lower learning rate tends to delay model collapse in our experiments --- though we see performance decay on AIME 2024 and 2025 within our training budget, and hypothesize that training for longer with \algo{}, even with a smaller learning rate, will lead to complete model collapse as usual.}
    \label{fig:learning_rate_all_datasets}
\end{figure}

Another common hyperparameter to tune is the learning rate. To investigate the effect learning rate has on \algo{}, we finetune a Qwen2.5-Math-7B model with different learning rates using \algo{} on the DAPO dataset, with all other hyperparameters kept fixed at their default values described in Appendix~\ref{app:training_hparam_details}. Figure~\ref{fig:learning_rate_all_datasets} shows our empirical findings: lowering learning rate seems to prevent model collapse within our training budget. However, we notice performance degradation on the harder AIME 2024 and 2025 datasets within our training budget, and hypothesize that prolonged training with \algo{}, even with a considerably lower learning rate, would still lead to model collapse. We could not study this in detail due to computational constraints, and leave this for future work.

\subsection{Different Number of generations per prompt}

\begin{figure}[h!]
    \centering
    \includegraphics[width=0.98\linewidth]{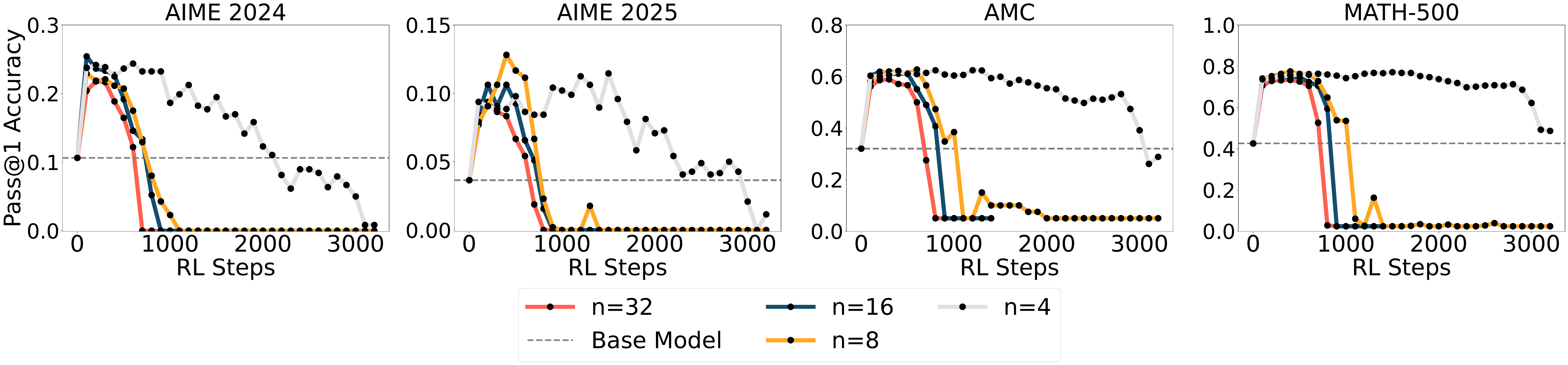}
    \caption{\footnotesize \textbf{(\algo{} with number of generations per prompt)} We compare the behavior of \algo{} with different number of generations/rollouts per prompt and report performance on all test datasets. All experiments here use a Qwen2.5-Math-7B model trained with RLOO on the DAPO dataset, with the hyperparameters other than number of generations per prompt kept fixed at the default ones described in Appendix~\ref{app:training_hparam_details}. Surprisingly, lowering the number of generations per prompt seems to delay model collapse.}
    \label{fig:num_gen_per_prompt_all_datasets}
\end{figure}

We observe the most surprising result among our hyperparameter tuning experiments when we vary the number of rollouts per prompt during training. Similarly as before, we finetune a Qwen2.5-Math-7B model with \algo{} on the DAPO dataset, with all hyperparameters (except number of rollouts per prompt) kept fixed at the their default values described in Appendix~\ref{app:training_hparam_details}. Figure~\ref{fig:num_gen_per_prompt_all_datasets} records our results: model collapse happens progressively later in the training run as we lower the number of rollouts per prompt. We hypothesize that generating fewer rollouts makes estimating the ``true" majority voting label for a single prompt more noisy. This noisy estimation then makes hacking the reward signal harder as well, thereby delaying model collapse. We have not been able to study this phenomenon in more detail due to computational constraints, and leave studying this for future work.

\newpage

\subsection{Entropy Coefficient}

\begin{figure}[h!]
    \centering
    \includegraphics[width=0.98\linewidth]{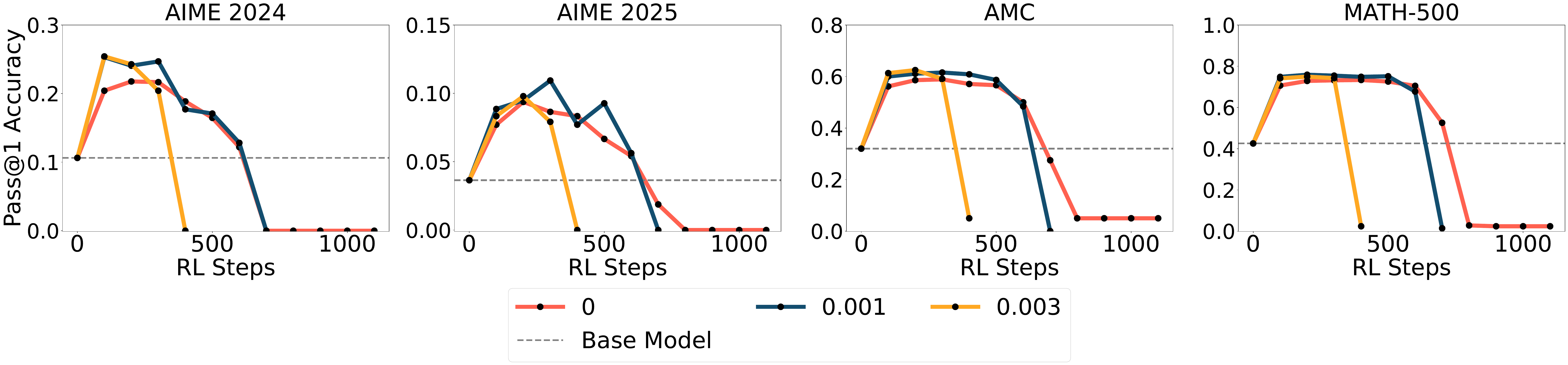}
    \caption{\footnotesize \textbf{(\algo{} with different entropy coefficient)} We compare the behavior of \algo{} with different entropy coefficients and report performance on all test datasets. All experiments here use a Qwen2.5-Math-7B model trained with RLOO on the DAPO dataset, with the hyperparameters other than entropy coefficient kept fixed at the default ones described in Appendix~\ref{app:training_hparam_details}. Surprisingly, increasing the entropy coefficient hastens model collapse.}
    \label{fig:entropy_coefficient_all_datasets}
\end{figure}

The final hyperparameter with which we experiment is adding an entropy loss to our regular RL objective. The modified RL objective is listed below:

\begin{equation}
    \mathcal{L}_\text{entropy-augmented}(\pi_\theta) = \mathcal{L}_\text{RL}(\pi_\theta) - \alpha \mathcal{H}(\pi_\theta)
\end{equation}

where $\mathcal{L}_\text{RL}(\pi_\theta)$ is the regular RL loss objective, $\alpha$ is the entropy coefficient, and $\mathcal{H}(\pi_\theta)$ is the per-token entropy averaged across all tokens in all rollouts. A reasonable hypothesis is that adding entropy to the loss objective will prevent model collapse~\citep{cheng2025reasoningexplorationentropyperspective} by discouraging the model to converge to one solution for every prompt. However, Figure~\ref{fig:entropy_coefficient_all_datasets} shows results in the contrary: adding an entropy term to the loss function and increasing the corresponding coefficient $\alpha$ accelerates model collapse. Upon inspection of the rollouts, we see that increasing $\alpha$ incentivizes the model to generate more random tokens in the rollouts, which maximizes entropy, followed by the same template final answer, which maximizes training (pseudo-)reward. This suggests that a better mechanism to prevent model diversity collapse is needed~\citep{song2025outcomebasedexplorationllmreasoning,zhou2025evolvinglanguagemodelslabels}, which we leave to future work to study.

\newpage

\section{Test-Time Self-Improvement}

\subsection{\algo{} can be used for test-time training}

\label{sec4.test.time.self.improvement}
\begin{figure}[h!]
    \centering
    \includegraphics[width=0.98\linewidth]{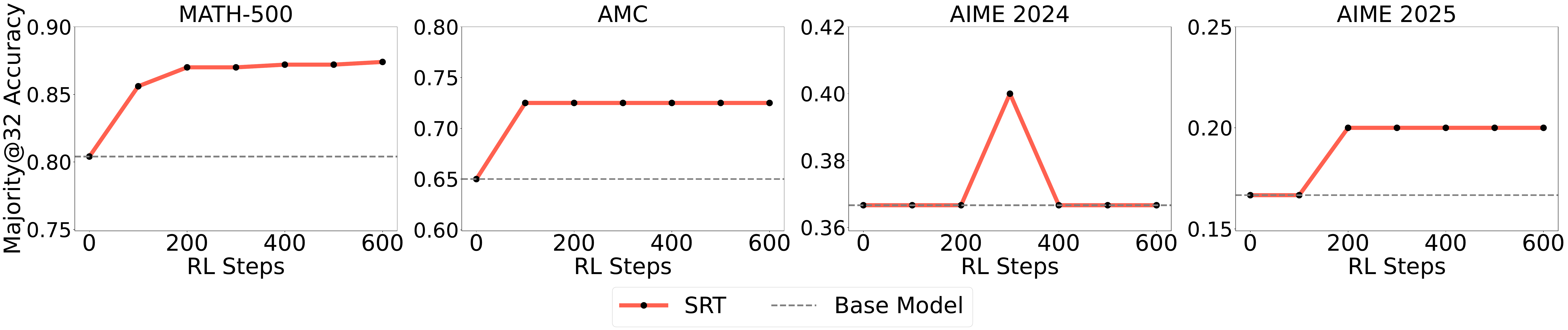}
    \caption{\footnotesize \textbf{(Test-Time Self-Training Performance)} Given the test dataset $\mathcal{D}_\text{test}$, one can perform \algo{} on $\mathcal{D}_\text{test}$ before making predictions. Our results show that this improves the majority voting performance on $\mathcal{D}_\text{test}$ without access to ground truth labels. Notice that the y axis is showing majority@32 accuracy instead of average@32 accuracy, for a fairer comparison with the baseline.}
    \label{fig:ttrl_main_paper}
    \vspace{-0.3cm}
\end{figure}

An appealing application of self-training is improving model accuracy via test-time training \citep{sun2020testtime, wang2021tent}, a direction also explored by the concurrent work of \citet{zuo2025ttrl}.
Test-time training refers to the procedure of further adapting or fine-tuning a pre-trained model on the actual test set itself, typically without access to labels or ground truth annotations. Applying \algo{} as a test-time training technique is remarkably straightforward: the unlabeled test set is treated precisely as if it were a training dataset, and \algo{} is directly applied. 

We compare the test-time performance of majority voting after SRT test-time training as well as without any test-time training.
Empirically, we observe (Fig \ref{fig:ttrl_main_paper}) that test-time training via \algo{} provides relatively limited, yet noticeable, performance gains when measured under the maj@32 metric, compared to the popular majority voting baseline applied directly to outputs generated by the base model. 

\subsection{Why Doesn’t the Performance Collapse during Test-Time-Training?}

\begin{figure}[h!]
    \centering
    \includegraphics[width=0.99\linewidth]{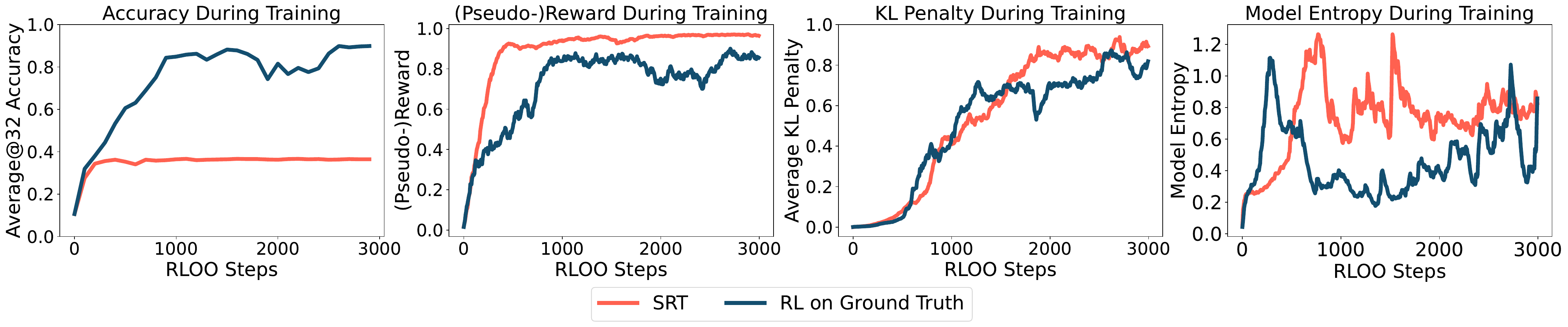}
    \caption{\footnotesize \textbf{(Test-Time Self-Training Dynamics)} We apply test-time training on AIME 2024 and observe no performance collapse. However, SRT's performance quickly saturates (leftmost plot), and the pseudo-reward value (second plot) also approaches saturation.} 
    \label{fig:aime_2024_ttrl_training_dynamics}
    \vspace{-0.3cm}
\end{figure}

Interestingly, upon completion of test-time training, a visual inspection of model's outputs reveals that the model's predictions still degenerate to a single response for nearly every test prompt---precisely the behavior identified as optimal solution to the \algo{} objective; however, the test-time accuracy remains high.

We conjecture that test-time self-training is inherently more stable due to crucial differences in dataset size. For example, consider the AIME24 test dataset, which contains only 30 samples for self-improvement. With such a limited sample size, the model quickly converges to a stable majority vote answer on these examples by reinforcing the particular chain-of-thought reasoning that leads to such solutions. After reaching this convergence, \algo{} ceases to receive meaningful gradient signals for further parameter updates, naturally stabilizing test-time performance (see     Figure \ref{fig:aime_2024_ttrl_training_dynamics} for test-time training dynamics).

In contrast, during regular training on large-scale datasets, the iterative supply of many fresh samples continually pushes the model to optimize heavily for consistency. In such conditions, the model is incentivized to adopt an overly simplistic generalization strategy (producing same \texttt{\textbackslash boxed\{\}} answer)—eventually collapsing by producing a uniform, prompt-independent prediction.

\section{More on Model Collapse Resulting from Self-Training}
\label{sec5.mitigation}

As discussed before, the optimization objective in \algo{} can lead to significant initial improvements followed by eventual model collapse. 
Here, we explore complementary strategies to address model collapse and further enhance the performance achievable via self-training:
\begin{enumerate}[label=\textbf{(\arabic*)}]
\item An \emph{early stopping} strategy leveraging a small labeled validation dataset to detect and prevent model collapse.
\item An \emph{algorithmic} strategy that \textit{delays} model collapse by using pseudo-labels generated from a stable base model rather than from the continuously updated model.
\item A \emph{data-driven} curriculum-based strategy to enhance model performance beyond simple early stopping.
\end{enumerate}

\subsection{Early Stopping}

\begin{figure}[h]
    \centering
    \includegraphics[width=0.7\linewidth]{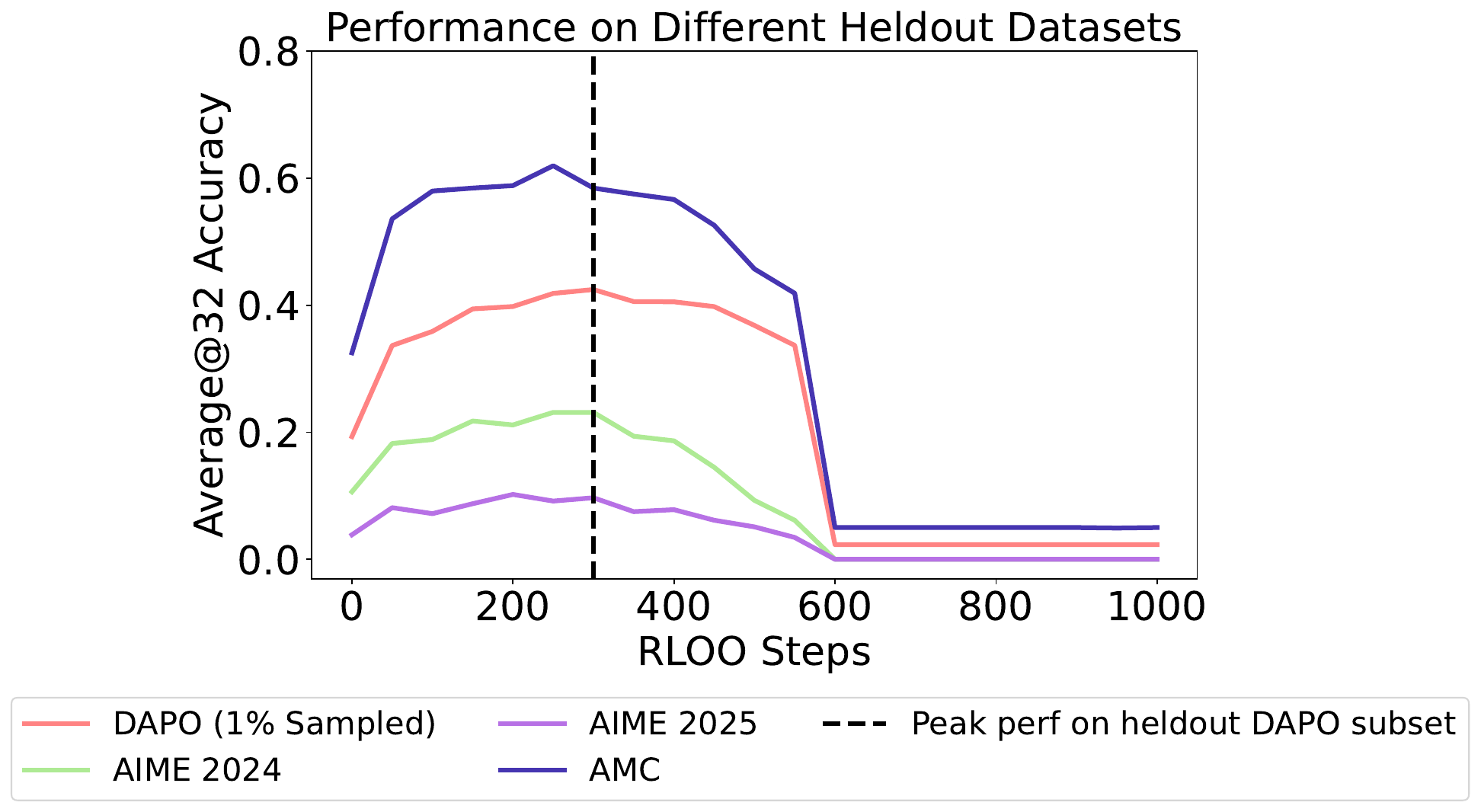}
    \caption{\footnotesize \textbf{(Early Stopping is Effective)} The peak performance occur at nearly the same point for all heldout sets, so using any would be effective for early stopping.}
    \label{fig:early_stopping}
\end{figure}

In our algorithm, even a small labeled validation dataset can effectively identify the peak performance point during self-training, thereby mitigating model collapse.
Figure~\ref{fig:early_stopping} shows the progression of model performance, measured throughout training on the DAPO dataset and evaluated across several test sets. Crucially, we find that the peak performance consistently occurs around the same training step across different held-out datasets. 
The vertical line in Figure~\ref{fig:early_stopping} marks early stopping using only 1\% of DAPO as validation, with performance on other datasets remaining near-optimal.

\subsection{Self-Training with Offline-Generated Labels}

\begin{figure}[h]
\centering
\includegraphics[width=0.8\linewidth]{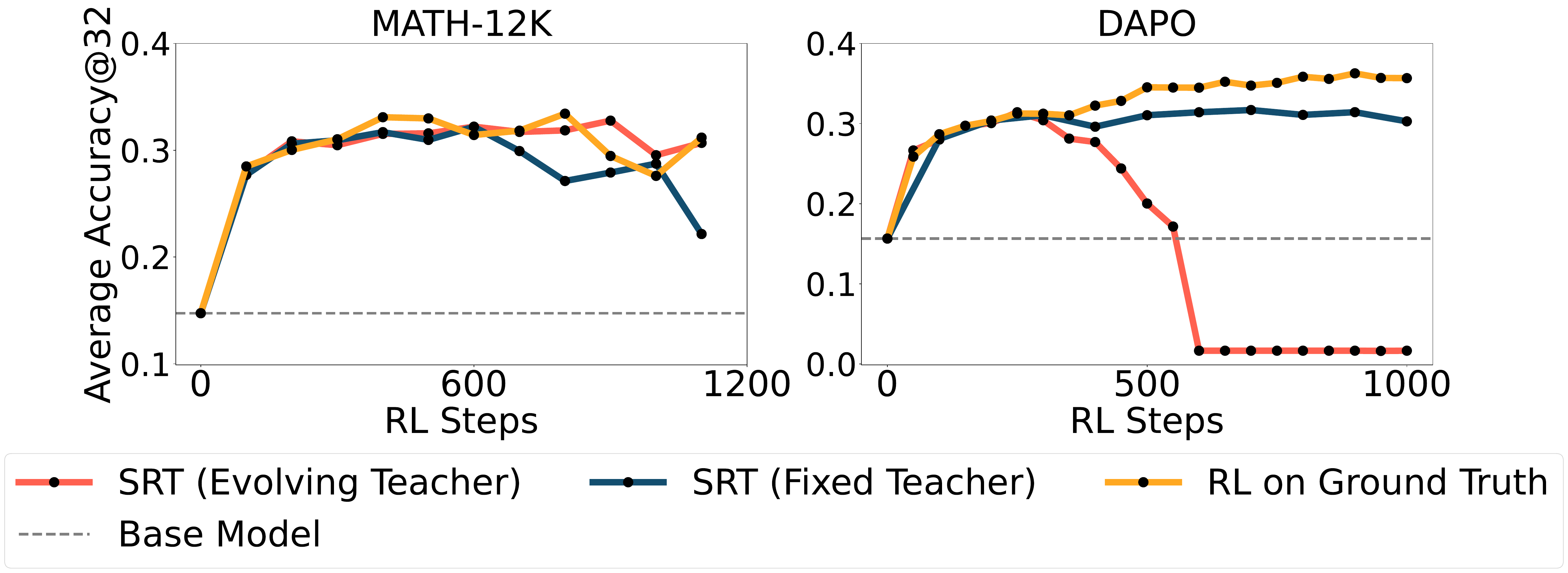}
\caption{\footnotesize \textbf{(Self-Training with Offline-Generated Labels)} Performance comparison between \algo{} and its variant, where pseudo-labels are precomputed from a fixed base model checkpoint. The fixed teacher labels maintain training stability while achieving comparable performance to \algo{}, highlighting a limitation of the online labeling strategy. Note that training starts to collapse for MATH-12k dataset ~1000 steps (which is roughtly ~1.5 epoch).}
\label{fig:training_on_offline_labels}
\end{figure}

The tendency toward model collapse arises because \algo{} prioritizes consistency over correctness, increasingly rewarding model agreement even if incorrect. A straightforward yet effective approach that \textit{delays} (but not completely prevent) the model collapse is to get the pseudo-labels from a stable, previously fixed checkpoint, rather than leveraging labels from the evolving policy (but with the downside that the generated labels will not benefit from the improvement in majority voting as training progressed, observed in Figure~\ref{fig:reasoning-gym-main-plot}). Here we generate pseudo-labels via majority voting rollouts from the Qwen2.5-Math-7B base model, store these offline-generated labels, and subsequently perform RL training against them.
Figure~\ref{fig:training_on_offline_labels} demonstrates that training with these offline-generated labels significantly stabilizes training while achieving performance comparable to \algo{}. Importantly, this indicates that the dynamic updating of pseudo-labels during training (online labeling) may not always confer substantial benefits, and instead can contribute to training instability. However, we \textbf{noticed training collapse even with offline labeled datasets when trained for longer than one epoch} for the MATH-12k dataset. 

\subsection{Self-Training with Curriculum Learning} \label{sec:self_training_with_curriculum}

\begin{figure}[h!]
\centering
\includegraphics[width=0.9\linewidth]{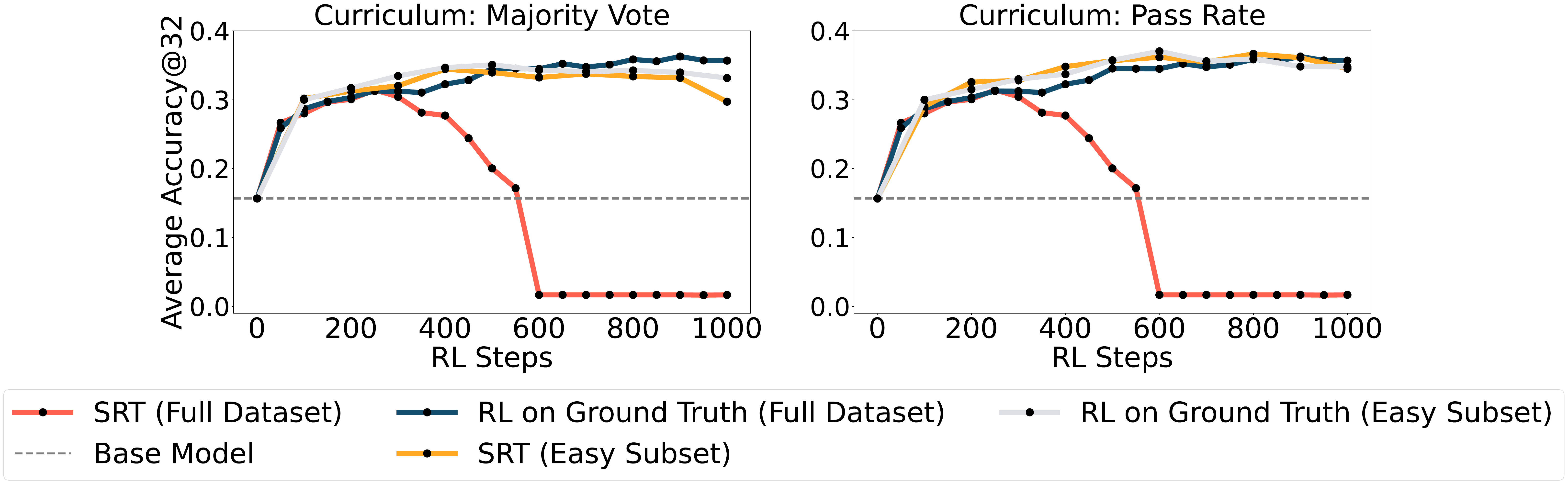}
\caption{\footnotesize \textbf{(Curriculum-Based Self-Training)} Performance of \algo{} on curated subsets containing the easiest 1/3 of prompts from the DAPO dataset, selected based either on model pass rate or frequency of the majority vote. Training on these easier subsets prevents reward hacking even after extensive training (3 epochs), demonstrating the effectiveness of curriculum learning strategies in sustaining continual model improvement.}
\label{fig:majority_voting_on_easy_subset}
\end{figure}
Our third approach, curriculum learning, is motivated by the observation that the model experiences earlier collapse when training on the difficult DAPO dataset compared to the simpler MATH-12K dataset. 
The intuition is that, on a more challenging dataset, the model finds it easier to abandon its pretrained knowledge in favor of optimizing self-consistency rather than genuinely learning to solve the underlying task.

We leverage this hypothesis to implement a curriculum learning strategy~\citep{bengio2009curriculum,andrychowicz2017hindsight,portelas2020automatic,florensa2017reverse,song2025mind, lee2025selfimprovingtransformersovercomeeasytohard,tajwar2025traininggenerallycuriousagent} by identifying the `easiest' subset of the DAPO dataset. To be precise, we retain 1/3-rd of the easiest DAPO prompts selected according to two distinct metrics:
\begin{enumerate}[label=(\arabic*), leftmargin=1cm]
\item \textbf{Pass rate of the base model}, which utilizes ground-truth labels.
\item \textbf{Frequency of the majority vote}, which does not require ground-truth labels.
\end{enumerate}
Figure~\ref{fig:majority_voting_on_easy_subset} shows that training on these easier subsets significantly delays the onset of reward hacking, allowing for continuous improvement even across multiple epochs. Remarkably, performance on these curriculum subsets reaches levels comparable to standard RL training with ground-truth labels on the entire DAPO dataset. More importantly, \textbf{we did not observe training collapse even after ~3 epochs of training on 1/3rd of "easy" DAPO} dataset. 
These promising results suggest that curriculum strategies may further extend the benefits of \algo{}, which we leave as a future research direction.

\subsection{Training Dynamics of \algo{} (Qwen2.5-Math-7B) on the Easy DAPO Subset}

\begin{figure}[h!]
    \centering
    \includegraphics[width=0.99\linewidth]{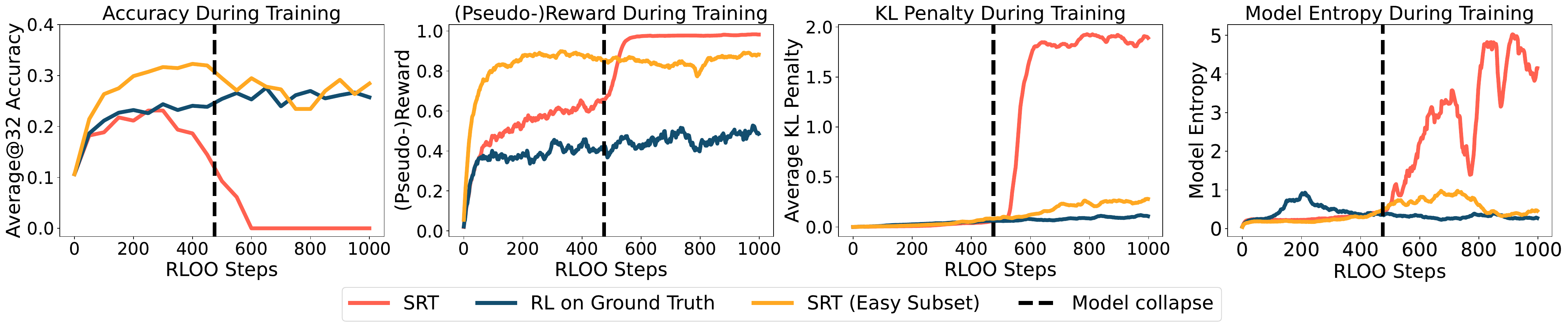}
    \caption{\footnotesize \textbf{(Training Dynamics of \algo{} on the Easy Subset of DAPO, using Qwen2.5-Math-7B)} We show the training dynamics of \algo{} on the easiest 1/3-rd of the DAPO dataset, chosen by ground truth pass rate of the base model. Compared to \algo{} on the entire DAPO dataset, \algo{} on the easier subset does not show any signs of reward hacking, even after taking 3 full passes over the training set. }
    \label{fig:easy_dapo_training_dynamics}
\end{figure}

Figure \ref{fig:dynamics_of_model_collapse} showed the common signs of reward hacking during \algo{}-training: namely, sudden drop in accuracy on a held-out dataset, sudden increase in KL penalty, etc. However, we found a simple yet effective way of mitigating reward hacking --- simply train on the easiest subset of the training data seems to retain the performance improvement obtained by training on the entire dataset, while preventing reward hacking within the same compute budget. Here we attempt to analyze this phenomenon further, from the lense of the same metrics we recorded in Figure \ref{fig:dynamics_of_model_collapse}.

Figure \ref{fig:easy_dapo_training_dynamics} shows our results on Qwen2.5-Math-7B: \algo{}-training on the easiest subset does not show the same behavior as training on the full dataset: accuracy on the heldout set does not drop, and KL penalty, while being slightly higher than that of training with ground truth, is still significantly lower than \algo{}-training on the full dataset. We also see that model entropy does not explode, so the model keeps outputting reasonable responses instead of the degenerate ones resulting from full dataset training. The most intriguing observation is that regarding pseudo-reward (Figure \ref{fig:easy_dapo_training_dynamics}, second from left): it very quickly gets very close to 1 and stabilizes around 0.9. This tends to suggest the model gets very little learning signal as the mean of the pseudo reward is already approximately 1, which is probably the reason it does not learn to reward hack within the same compute budget. We leave investigating this further for future work.

\subsection{Training Qwen3-14B-Base on the Easy DAPO Subset}

\begin{figure}[h!]
    \centering
    \includegraphics[width=0.99\linewidth]{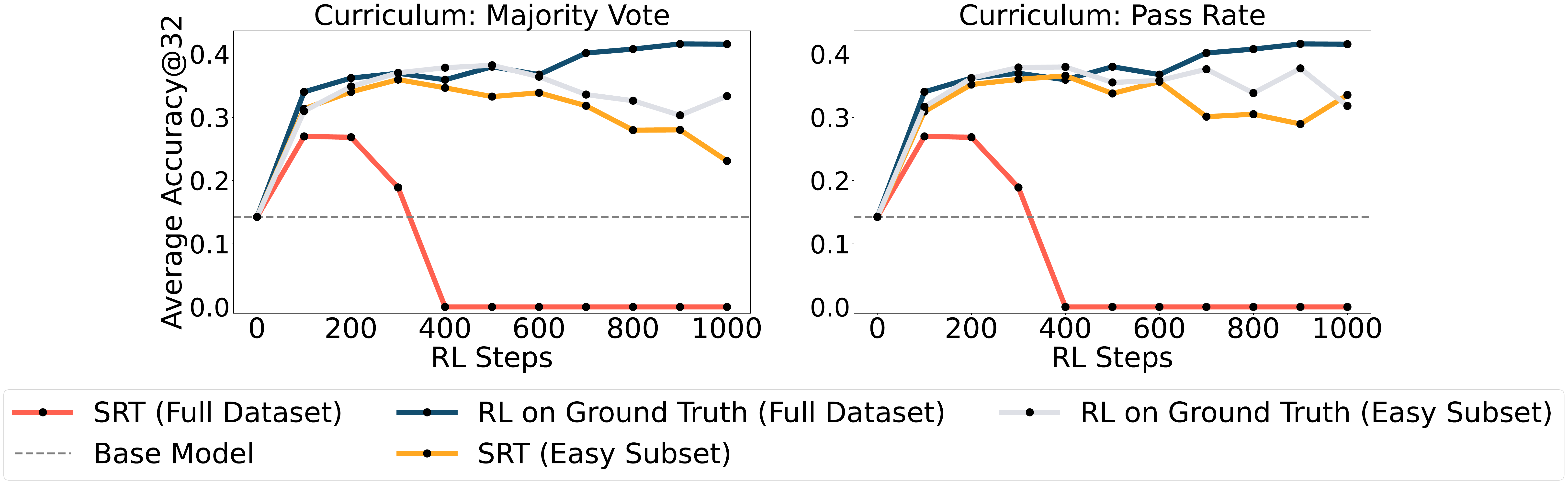}
    \caption{\footnotesize \textbf{(Qwen3-14B-Base Trained on the Easy DAPO Subset)} We take the same easy subsets of DAPO used in Figure~\ref{fig:easy_dapo_training_dynamics}  and train a Qwen3-14B-Base model with \algo{} on it. We see the same behavior as Qwen2.5-Math-7B (Figure \ref{fig:majority_voting_on_easy_subset}), that \algo{} on the easy subset does not exhibit performance collapse within the same compute budget.}
    \label{fig:qwen3_easy_dapo_subset}
\end{figure}

One of our most interesting observations is that simply using the easiest 1/3-rd of the DAPO dataset eliminates the performance collapse within our training budget (it can still happen if one trains more, though we do not observe it). We want to test whether this is still true for a different base model. To do so, we take the same easiest subset used in Figure~\ref{fig:easy_dapo_training_dynamics} (so the subset is determined using either the ground truth pass rate or the frequency of the majority answer of a Qwen2.5-Math-7B model) and train a Qwen3-14B-Base model with \algo{} on this subset. Figure \ref{fig:qwen3_easy_dapo_subset} shows the result of our experiments: similar to Qwen2.5-Math-7B model, the Qwen3-14B-Base model also does not exhibit performance collapse within the same training budget.

\subsection{Generating Easy DAPO Subset using Qwen2.5-Math-1.5B}

\begin{figure}[h!]
    \centering
    \includegraphics[width=0.99\linewidth]{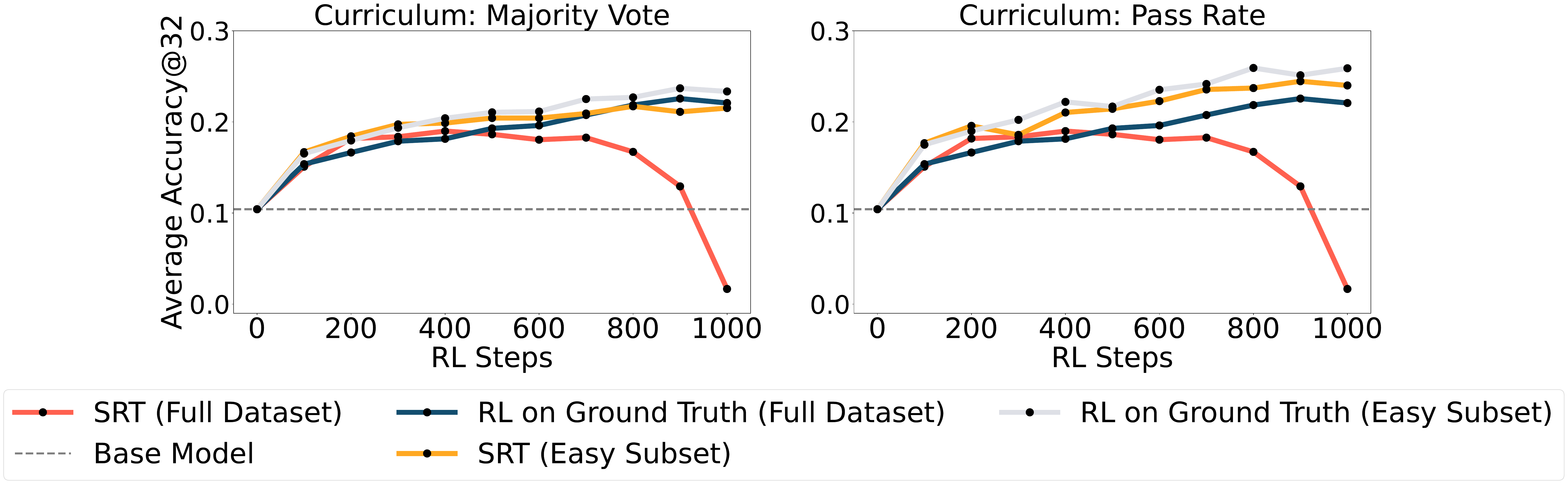}
    \caption{\footnotesize \textbf{(Generating Easy DAPO Subset Using Qwen2.5-Math-1.5B)} To see if the curriculum generation process is reproducible, we generate the easy 1/3-rd subset (by both majority voting frequency and pass rate) using a Qwen2.5-Math-1.5B model. This is in contrast with Figure \ref{fig:majority_voting_on_easy_subset} and Figure \ref{fig:qwen3_easy_dapo_subset}, where we used a Qwen2.5-Math-7B model for the easy subset generation. Furtheremore, we train the Qwen2.5-Math-1.5B on the easy subset as before, and make similar observations: training with \algo{} on a easy subset, even for 3 epochs, does not lead to performance collapse. }
    \label{fig:qwen2.5_math_1.5B_easy_dapo_subset}
\end{figure}

Next, we want to test if the easy subset generation process of our curriculum algorithm itself is reproducible using different base models. To do so, we generate the easy 1/3-rd subset using a Qwen2.5-Math-1.5B model by both majority voting frequency and pass rate. This is in contrast with our earlier sections, especially Figure \ref{fig:majority_voting_on_easy_subset} and Figure \ref{fig:qwen3_easy_dapo_subset}, where we used a Qwen2.5-Math-7B model for the subset generation. We also train the Qwen2.5-Math-1.5B model on the resulting easy subsets. Figure \ref{fig:qwen2.5_math_1.5B_easy_dapo_subset} shows our results --- we see the same trend as our earlier result, i.e., training on the easy subsets, even for 3 epochs, does not lead to any performance collapse. Moreover, surprisingly, up to our training budget, \algo{} on the easy subset matches the performance of RL training with ground truth labels.

\subsection{\algo{} training with online curriculum}

\begin{figure}[h!]
    \centering
    \includegraphics[width=0.99\linewidth]{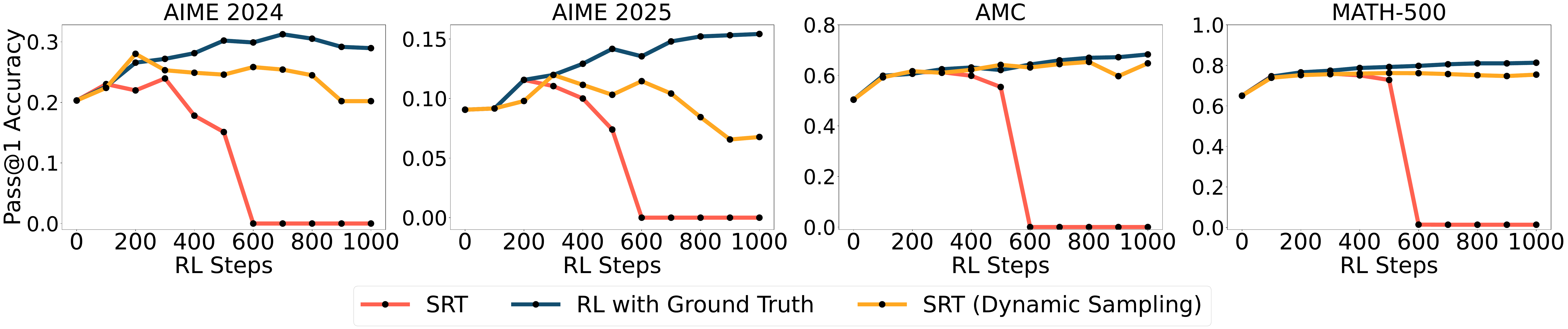}
    \caption{\footnotesize \textbf{(Dynamic oversampling then filtering delays model collapse within training budget)} Here we experiment with the idea that dynamic oversampling then filtering prompts with low model self-consistency can prevent/delay model collapse by improving the training signal. All experiments are done using Qwen2.5-Math-7B trained on the DAPO dataset using hyperparameters described in Appendix~\ref{app:training_hparam_details}. For the dynamic sampling variant of \algo{}, we discard prompts where the majority voting answer does not occur at least 60\% of all 32 generations for the prompt, and keep generating until we have at least 16 such prompts (the prompt batch size used for all baselines). We see that this dynamic oversampling then filtering variant of \algo{} can prevent model collapse within our training budget of 1000 RL steps, though prolonged training can still lead to model collapse in laters phases of the training.}
    \label{fig:dynamic_oversampling}
\end{figure}

Based on the insights we gained from the previous sections, the final strategy we experiments with to mitigate model collapse with \algo{} is dynamic oversampling then filtering the prompts, similar to the one introduced by DAPO~\citep{yu2025dapoopensourcellmreinforcement}. Concretely, let the prompt batch size be $B$. Then we employ the following routine: \textbf{(1)} Sample $B$ prompts from the training dataset, \textbf{(2)} Generate $K$ responses per prompt, calculate per prompt majority votes and the fraction of generations per prompt where the final answer is the same as the majority vote (i.e., self-consistency level), \textbf{(3)} Discard prompts where the majority voting fraction is smaller than a threshold, and repeat the prior steps until we have $B$ prompts and their corresponding generations, \textbf{(4)} Take one RL step on the generated dataset.

We investigate this idea with a Qwen2.5-Math-7B model trained on the DAPO dataset. We set the filtering threshold to be $0.6$ (so any prompt where the majority vote appears in less than 60\% of all generations is discarded), and all other hyperparameters are kept fixed at the default ones from Appendix~\ref{app:training_hparam_details}. Figure~\ref{fig:dynamic_oversampling} shows our empirical findings: we see that our intuition holds, and this dynamic oversampling and filtering procedure prevents model collapse within our training budget. We hypothesize that extended \algo{}-training would still lead to model collapse since the reward hacking solution is still an optimal solution under our training objective and it is still possible for the model to discover it during training. Despite this, our procedure of controlling the task distribution is effectively filtering away the harder prompts dynamically during training and employing an online curriculum, which results in delaying model collapse, just as we expect from our experiments using only offline curriculum for \algo{}-training in Appendix~\ref{sec:self_training_with_curriculum}.

\newpage

\section{Detailed Experiment Results using non-Qwen models}

To validate the efficacy of \algo{} on LLMs with different pre-training/post-training routine, we run additional experiments on two more models: Deepseek-Math-7B-Instruct~\citep{shao2024deepseekmathpushinglimitsmathematical} and Llama-3.1-8B-Instruct~\citep{grattafiori2024llama3herdmodels}. Our results are described below.

\subsection{Deepseek-Math-7B-Instruct}

\begin{figure}[h!]
    \centering
    \includegraphics[width=0.99\linewidth]{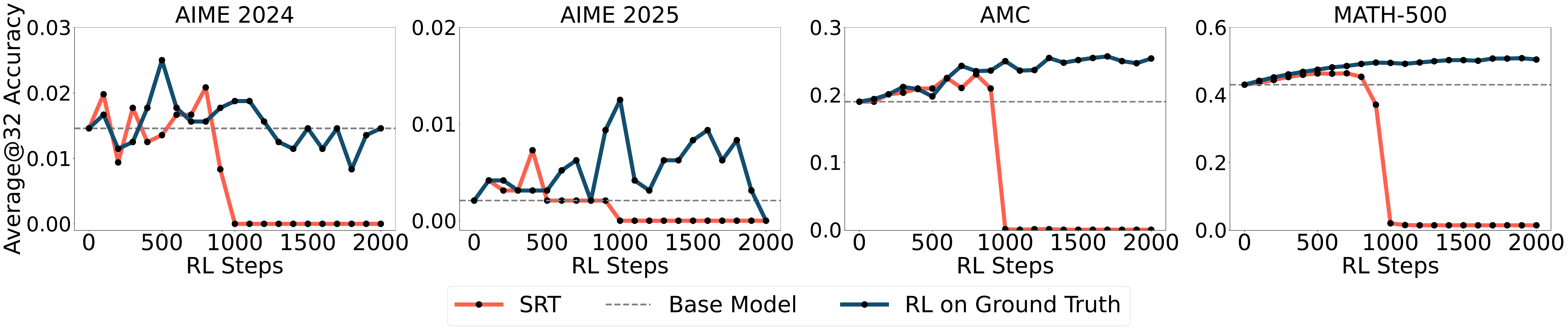}
    \caption{\footnotesize \textbf{(Training Deepseek-Math-7B-Instruct on MATH-12K using \algo{})} We see similar trends for \algo{}-training on Deepseek-Math-7B-Instruct as we saw on our experiments with Qwen models: \algo{} initially matches performance gain obtained with RL training with ground truth, but faces performance collapse after prolonged training.}
    \label{fig:deepseek_on_math_12k}
\end{figure}

We train Deepseek-Math-7B-Instruct on the MATH-12K dataset, and test on AIME 24, AIME 2025, AMC and MATH-500. For training, we use the same hyperparameters use used to train Qwen2.5-Math-7B-Instruct due to lack of compute for running a sweep over possible hyperparameters. We note that we did not find the recommended temperature or other sampling parameters for the Instruct model in~\citep{shao2024deepseekmathpushinglimitsmathematical}, but their base models were evaluated with temperature 0.7, so we choose temperature 0.7, top-p 1.0 and no top-k sampling for our evaluations. Figure~\ref{fig:deepseek_on_math_12k} shows our results: we see similar trends as our earlier experiments, where \algo{} initially matches performance gain from RL with ground truth, but then leads to performance collapse after prolonged training. 

\subsection{Llama-3.1-8B-Instruct}

\begin{figure}[h!]
    \centering
    \includegraphics[width=\linewidth]{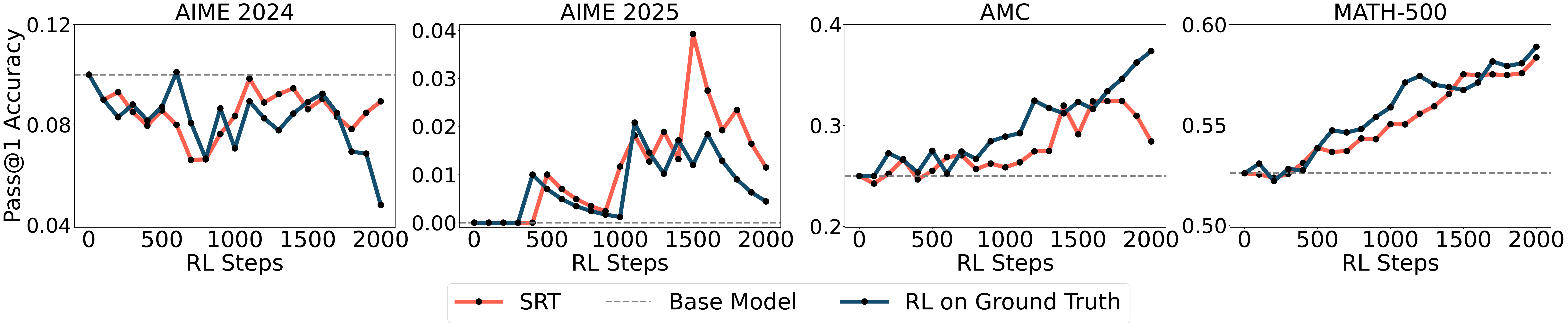}
    \caption{\footnotesize \textbf{(Training Llama-3.1-8B-Instruct on Big-Math-RL-Verified with learning rate $10^{-7}$)} Llama-3.1-8B-Instruct, when trained on a filter subset of the Big-Math dataset, shows significant gains on MATH-500 from both \algo{} and RL with ground truth. In fact, up to our training budget of 2K steps, both seem to improve performance at the same rate, from 52.6\% to around 60\%.}
    \label{fig:llama_8b_on_big_math}
\end{figure}

\begin{figure}[h!]
    \centering
    \includegraphics[width=\linewidth]{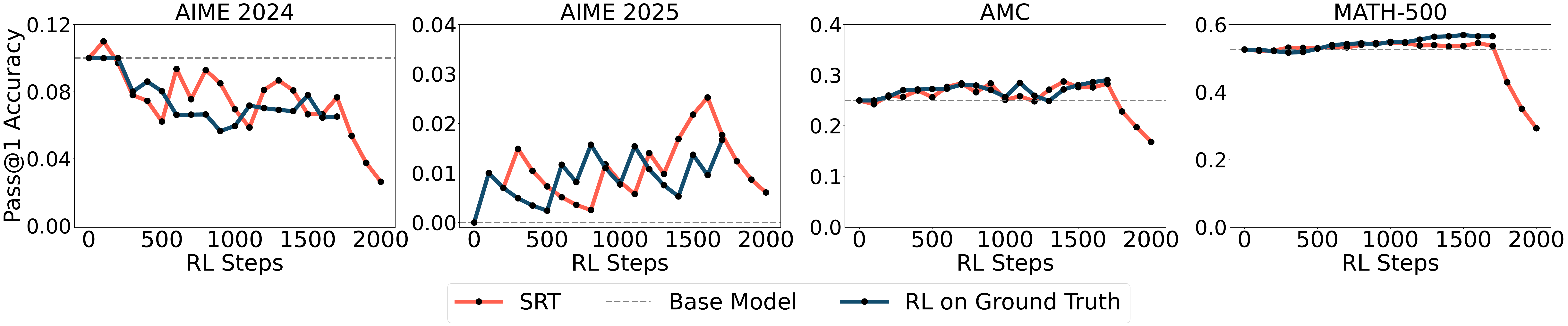}
    \caption{\footnotesize \textbf{(Training Llama-3.1-8B-Instruct on Big-Math-RL-Verified with learning rate $3 \times 10^{-7}$)} Llama-3.1-8B-Instruct, when trained on a filter subset of the Big-Math dataset with a higher learning rate ($3 \times 10^{-7}$ instead of $10^{-7}$ in Figure~\ref{fig:llama_8b_on_big_math}) demonstrates model collapse within the same training budget.}
    \label{fig:llama_8b_on_big_math_higher_lr}
\end{figure}

Llama-3.1-8B-Instruct showed no gains while being trained on DAPO or MATH-12K. This can be due to insufficient hyperparameter tuning or the model's starting performance on these datasets not suitable for learning. So we chose the Big-Math dataset~\citep{albalak2025bigmathlargescalehighqualitymath}, a dataset with over 250,000 math questions with verifiable answers. Moreover, this dataset has been constructed by filtering common evaluation datasets like MATH-500, making it suitable for our purposes. \textbf{The primary benefit of using this dataset is that it comes with Llama-3.1-8B-Instruct pass rate}, so we can easily ascertain the difficulty of each datapoint and aggregate a subset that can be suitable for training the Llama model. Specifically, we take the subset of Big-Math where Llama-3.1-8B-Instruct has pass rate between 0.3 and 0.7, to filter away too easy or too difficult questions. Next, we train the model on this subset using the same hyperparameters as we used for training the Qwen2.5-Math-7B-Instruct, except we lower the learning rate to $10^{-7}$ (from $10^{-6}$), as we see that leads to more stable learning curves. For evaluation, we use the same prompt template but temperature 0 (greedy decoding), to match the starting model's performance reported on its model card, and report pass@1 accuracy.

Figure~\ref{fig:llama_8b_on_big_math} shows our results: on MATH-500, Llama-3.1-8B-Instruct shows the same performance growth when trained via \algo{} or RL with ground truth, up to our training budget of 2000 steps. Performance growth is also significant, and training improves pass@1 accuracy from 52.6\% to around 60\%. Note that we do note report performance on the harder datasets like AIME, because the Llama model's performance remain close to 0 throughtout training (with both objectives) on these datasets, signalling that they might be too hard for this model. We also ran an additional experiment using a higher learning rate of $3 \times 10^{-7}$. Figure~\ref{fig:llama_8b_on_big_math_higher_lr} shows our empirical findings: with the higher learning rate, Llama-3.1-8B-Instruct also start to show model collapse within our training budget.

\newpage

\section{Example Tasks from Reasoning Gym} \label{app:reasoning_gym_example_tasks}

\begin{tcolorbox}[examplebox, title=Example Tasks]
\textbf{Task: Family Relationships (Level 4)}\\

\emph{Question:} John is married to Isabella. They have a child called Edward. Edward is married to Victoria.

What is Isabella to Edward? Respond only with the word that describes their relationship.\\

\emph{Answer:} mother

\vspace{0.8em}

\textbf{Task: Bitwise Arithmetic (Level 2)}\\

\emph{Question:} Please solve this problem. Assume there is arbitrary bit depth and that there are signed integers. If the answer is negative, reply as a negative value (e.g., \(-0x3\)), not the two's-complement form. Reply only with the final hexadecimal value.\\[0.25em]
\[
((0x3a24 - 0x24b8) + (0x1741 \gg 0x3))
\]\\[-0.25em]

\emph{Answer:} 0x1854

\vspace{0.8em}

\textbf{Task: Knights and Knaves (Level 2)}\\

\emph{Question:} A very special island is inhabited only by sages and fools. Sages always tell the truth, and fools always lie. You meet 2 inhabitants: Zoey and Riley. Zoey commented, ``Riley is a fool.'' In Riley's words: ``Zoey is a sage or Riley is a sage.'' So who is a sage and who is a fool? (Format your answer like: ``Zoey is a sage/fool, and Riley is a sage/fool'')\\

\emph{Answer:} Zoey is a fool, and Riley is a sage.
\end{tcolorbox}

Above of we see one example from each of the three Reasoning Gym tasks used in our work. The examples shown are of the lowest difficulties that we first train the model on using ground truth RL. We do self training on more difficult variants on the tasks. Difficulty can be changed by the modifying the either number of person or digits. In this work, we abstract it away by calling it "level".

\newpage

\section{Behavior of a Model after Performance Collapse}\label{app:reward_hacking_examples}

We provide example generations from a Qwen2.5-Math-7B model before and after prolonged training with \algo{} on DAPO in Table \ref{tab:reward_hacking_examples}. After \algo{} training for 1200 steps, the model exhibits performance collapse and generates  \texttt{\string\boxed\{1\}} together with some other incoherent set of tokens for all prompts. This also shows that the model has learned to reward hack the \algo{} objective, since outputting the same answer for all queries will invariably lead to maximum consistency regardless of the accuracy of the answer.

\begin{table*}[h]
    \caption{\footnotesize Three examples of model responses for the same prompt, before and after prolonged training with \algo{} on the DAPO dataset, for a Qwen2.5-Math-7B model. Notice that for some prompts, the model responses before training ends before completion, this is due to the model running out of our token generation budget. The model after 1200 steps of \algo{} training exhibits performance collapse, and it outputs \texttt{\string\boxed\{1\}} and some other incoherent set of tokens irrespective of the given prompt.}
    \vspace{0.2cm}
    \label{tab:reward_hacking_examples}
    \centering
    \resizebox{\textwidth}{!}{
        \begin{tabular}{|p{4cm} |p{6cm} | p{6cm}|}
        \toprule
        Prompt & Response before any training & Response after 1200 steps of \algo{} training \\
        \hline
        How many positive perfect squares less than $2023$ are divisible by $5$? & To determine how many positive perfect squares less than 2023 are divisible by 5, let's break the problem down.
        1. Identify the range of perfect squares: We need to find all perfect squares \(n^2\) such that \(n^2 < 2023\). This means \(n < \sqrt{2023}\). Since \(\sqrt{2023} \approx 44.98\), the largest integer \(n\) for which \(n^2 < 2023\) is 44.
        2. Check for divisibility by 5: For a perfect square \(n^2\) to be divisible by 5, \(n\) itself must be divisible by 5. So, we need to f & espionage----- \texttt{\string\boxed\{1\}} \\
        \hline
         What is the value of
         \[2^3 - 1^3 + 4^3 - 3^3 + 6^3 - 5^3 + \] \[\dots + 18^3 - 17^3?\] &
         What is the final answer within \texttt{\string\boxed\{\}}.
        
        \# Define the range of numbers and the expression to calculate ... \textcolor{red}{(Model writes code here, we ignore for the sake of simplicity)}
        
        4046

        The value of the expression \(2^3 - 1^3 + 4^3 - 3^3 + 6^3 - 5^3 + \dots + 18^3 - 17^3\) is \texttt{\string\boxed\{4046\}}. & drained , \texttt{\string\boxed\{1\}} Zac MemoryStream \\
        \hline
            Find the sum of all integer bases $b>9$ for which $17_{b}$ is a divisor of $97_{b}$. & It is important to first convert the numbers given in base \( b \) to base 10 for mathematical operations.
    The number \( 17_b \) is \( 1 \cdot b + 7 \) in base 10.
    The number \( 97_b \) is \( 9 \cdot b + 7 \) in base 10.
    We need to find integer bases \( b > 9 \) such that \( 17_b \) divides \( 97_b \). In other words: \[ 9b + 7 = k (1b + 7) \] for some integer \( k \).
    Let's proceed with the necessary calculations in Python to ensure accuracy.
    \# define a functi & -//////\texttt{\string\boxed\{1\}} \\
    \bottomrule
        
    \end{tabular}
    }
\end{table*}

\end{document}